\documentclass{article}
\usepackage{preprint,times}

\usepackage{amsmath,amsfonts,bm}

\def\eqref#1{equation~\ref{#1}}

\def\1{\bm{1}}

\def\vx{{\bm{x}}}
\def\vy{{\bm{y}}}

\DeclareMathAlphabet{\mathsfit}{\encodingdefault}{\sfdefault}{m}{sl}
\SetMathAlphabet{\mathsfit}{bold}{\encodingdefault}{\sfdefault}{bx}{n}

\usepackage{hyperref}
\usepackage{url}
\usepackage{graphicx}
\usepackage{stmaryrd}
\usepackage{amsmath}
\usepackage{float}
\usepackage{subcaption}

\usepackage{booktabs}
\usepackage{amsfonts}
\usepackage{nicefrac}
\usepackage{microtype}
\usepackage{xcolor}

\title{Can AI Forecast the Extremes?\\ Evaluating Probabilistic Weather Models\\ on Tropical Cyclones}

\author{Nathan Chalumeau\\
INRIA, Paris, France\\
AXA, Paris, France \\
\texttt{nathan.chalumeau@inria.fr} 
\And
Xavier Renard\\
Sorbonne Université, CNRS, LIP6, F-75005 Paris, France\\
AXA, Paris, France\\
\texttt{xavier.renard@axa.com} 
\And
Remi Meynadier\\
AXA, Paris, France\\
\texttt{remi.meynadier@axa.com}
\And
Hugo Rakotoarimanga\\
AXA, Paris, France\\
\texttt{hugo.rakotoarimanga@axa.com}
\And
Marcin Detyniecki\\
Sorbonne Université, CNRS, LIP6, F-75005 Paris, France\\
AXA, Paris, France\\
Polish Academy of Science, IBS PAN, Warsaw, Poland \\
\texttt{marcin.detyniecki@axa.com} 
\And
 Emmanuel de Bezenac\\
 INRIA, Paris, France\\
 \texttt{emmanuel.de-bezenac@inria.fr} 
}

\begin{document}

\maketitle
\lhead{} 
\renewcommand{\headrulewidth}{0pt}

\begin{abstract}
Weather prediction is undergoing an important shift, with data-driven AI models outperforming leading numerical weather prediction systems on average at a fraction of the computational cost. Yet, whether this success extends to extreme events remains contested. Earlier evaluations have exposed important shortcomings, but have largely focused on deterministic models and metrics, leaving the capabilities of the latest probabilistic models less understood. We conduct an extensive benchmark of tropical cyclone forecasting, comparing GenCast, FourCastNet 3, and FGN with ECMWF's operational physics-based ensemble, IFS-ENS. We develop a suite of probabilistic metrics and diagnostic analyses to assess cyclone tracking, intensity biases, uncertainty, and the wind--pressure relationship that characterizes storm structure. We find that, despite remaining limitations in intensity and ensemble spread, leading AI models substantially outperform IFS-ENS in cyclone tracking: FGN achieves better probabilistic location scores for 90\% of evaluated storms in 2025. Further analyses reveal model-specific failure modes and suggest a lingering influence of pretraining data, pointing to opportunities for targeted improvement. Nonetheless, these results suggest that AI models are capable of capturing physical relationships that remain predictive well beyond typical weather conditions, motivating further investigation into how these capabilities emerge and how to strengthen them.
\end{abstract}

\section{Introduction}

Weather forecasting has long relied on numerical simulation of the atmosphere. Learned models now compete with leading numerical systems at a fraction of the simulation cost \citep{bi_pangu-weather_2022, lam_graphcast_2023}. But their success on standard benchmarks leaves a consequential question unresolved: how well do they predict rare events for which accurate forecasts matter most?

Their reliability at representing extremes remains a source of skepticism. Evaluations have exposed substantial shortcomings in forecasting record-breaking weather and severe tropical cyclones, raising concerns that the success of learned models depends on conditions well represented in their training data \citep{zhang_numerical_2025, sun_can_2025}. These findings have challenged the broader promise of AI weather forecasting. Yet, much of the evidence has focused on deterministic models. The recent progress of probabilistic models \citep{price_gencast_2024, bonev_fourcastnet_nodate, alet_skillful_2025} motivates revisiting this assessment, and the examination of both their ability to forecast extreme events and the uncertainty represented by the ensembles they simulate.

Tropical cyclones are among the most damaging weather hazards globally \citep{wmo_atlas_2021}. Evacuation and impact preparedness depend on forecasts of tropical cyclone location and intensity several days ahead, including the uncertainty of those predictions, issued by weather agencies. Yet cyclones occupy only about 0.5\% of grid points in global climate simulations \citep{lacombe_improving_2023}, and the proportion of intense cyclones is projected to increase under continued global climate warming \citep{intergovernmental_panel_on_climate_change_ipcc_climate_2023, knutson_tropical_2020}. Their rarity and consequences make them an important testbed for learned forecast models, particularly as these models enter operational and cyclone forecasting trials \citep{lang_aifs_2025, deepmind_weatherlab_2025}.

We revisit the ability of AI models at forecasting extreme weather through a systematic evaluation of three recent leading probabilistic AI weather models: GenCast \citep{price_gencast_2024}, FourCastNet 3 (FCN3; \citealp{bonev_fourcastnet_nodate}), and FGN \citep{alet_skillful_2025}. Across 178 tropical cyclones in two evaluation settings, we intercompare these three models' ability to predict location and intensity with respect to reanalysis and best track data. In addition, we benchmark these models against IFS-ENS, ECMWF's numerical operational ensemble model, to assess how far recent progress extends to cyclone forecasting and where limitations remain. Our evaluation examines both forecast skill and the properties of the predicted storms, combining probabilistic track forecasts, intensity, ensemble calibration, and wind--pressure dependence. This allows us to assess both forecast accuracy and the models' ability to reproduce the physical relationships that characterize these extreme events.

Among our findings, operational AI models improve cyclone forecasts in several respects. FGN outperforms IFS-ENS on probabilistic track scores for about 90\% of the 41 retained storms in 2025, and its advantage extends to the spatial location of strong winds and low pressures. Both operational AI models show larger relative CRPS gains over IFS-ENS near cyclones than at global scale. Yet improvements in aggregate scores do not imply improvements in every aspect of the storm: GenCast outperforms FCN3 on cyclone-local CRPS despite a larger peak-wind bias, while FGN's better spatial scores coexist with intensity biases similar to GenCast's and persistent ensemble calibration errors. Finally, the operational AI models retain intensity distributions close to ERA5 despite fine-tuning on HRES-fc0, yet reproduce HRES-fc0's wind--pressure dependence once marginal differences are removed. Together, these results distinguish substantial progress in predicting storm location and spatial structure from unresolved errors in intensity and uncertainty.

\section{Experimental setup}
\label{sec:setup}

\begin{figure}[H]
\centering
\includegraphics[width=\linewidth]{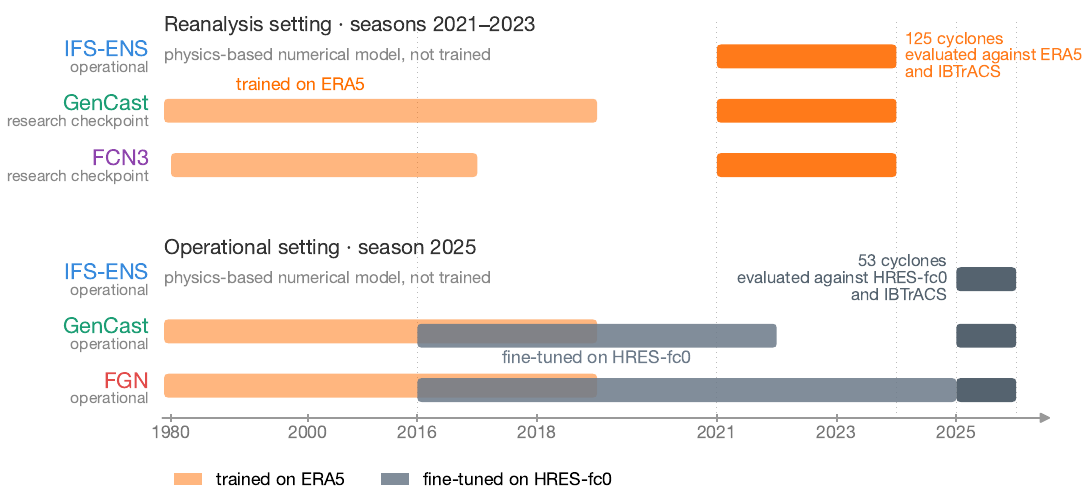}
\caption{\textbf{Evaluation windows of the benchmark.} To evaluate probabilistic models on tropical cyclones, evaluation years are chosen according to each model's training and fine-tuning periods.}
\label{fig:timeline}
\end{figure}

\paragraph{Models.}
We evaluate three leading probabilistic weather models that differ both in their training objectives and in the way they represent uncertainty.
\textbf{GenCast} uses conditional diffusion to sample future atmospheric states \citep{price_gencast_2024}.
\textbf{FourCastNet~3 (FCN3)} combines spherical convolutions with stochastic perturbations. It is trained using the continuous ranked probability score (CRPS) in the physical and spectral space \citep{bonev_fourcastnet_nodate}. 
\textbf{FGN} optimizes the CRPS at individual locations and forms an ensemble by combining four independently trained networks; within each network, additional members are generated via learned perturbations \citep{alet_skillful_2025}.

\paragraph{Evaluation settings.}

Four datasets serve as reference: \textbf{IBTrACS} \citep{knapp2010international} supplies observed tracks and intensities; \textbf{IFS-ENS} \citep{ecmwf_ifs_2024}, ECMWF's operational physics-based ensemble, is the numerical baseline, especially for uncertainty representation; \textbf{ERA5} \citep{hersbach_era5_2020} and \textbf{HRES-fc0} are the gridded references for two experimental configurations, with two versions of GenCast available. The \textbf{\textit{reanalysis setting}} uses GenCast and FCN3 trained exclusively on ERA5, evaluated against ERA5 and IBTrACS. The \textbf{\textit{operational setting}} uses GenCast (operational) and FGN, trained on ERA5 and fine-tuned on HRES-fc0, the initial conditions of ECMWF's high-resolution forecast system, evaluated against HRES-fc0 and IBTrACS. This design allows for the investigation of how the choice of reference data influences intensity bias, in particular whether ERA5-related biases remain after operational fine-tuning.
The evaluation years follow the model training periods (Figure~\ref{fig:timeline}), extending the tests over several cyclone seasons. Because the storms and model versions differ between settings, the comparison does not isolate the effect of fine-tuning.

\paragraph{Forecast protocol.}
We conduct a retrospective evaluation of historical tropical cyclones across all basins. 178 IBTrACS cyclones whose maximum reported one-minute sustained wind reaches at least 64 knots are selected and reproduced using the AI models. We assess predicted tracks, intensity, ensemble calibration, and storm structure, introducing each diagnostic alongside its scientific question. For each storm, we initiate a 15-day forecast rollout with 50 ensemble members per model in the reanalysis setting and 49 in the operational setting (Appendix~\ref{app:reproducibility}). Forecasts are initiated 12 hours before the first reported wind observation in the IBTrACS dataset, adjusted to the preceding forecast cycle where necessary. In total, the evaluation covers 26,541 member forecasts across lead times, using both cyclone tracks and atmospheric fields. Operational forecasts were obtained from existing archives, and we generated the forecasts for the reanalysis setting using more than 6,000 H100 GPU-hours. Appendix~\ref{app:reproducibility} details the initialization rules, forecast sources, matching procedures, and score aggregation.

\begin{figure}[H]
    \centering
    \includegraphics[width=\linewidth]{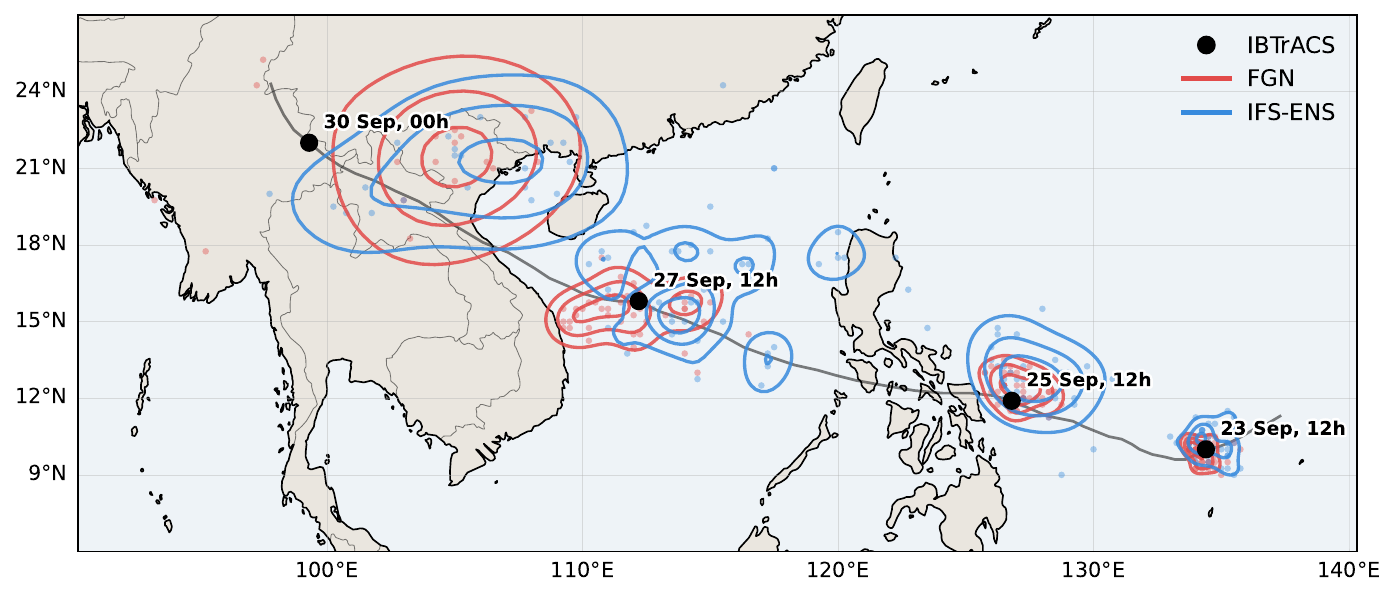}
    \caption{\textbf{Typhoon Bualoi (2025).} FGN (red) and IFS-ENS (blue) forecasts at four lead times: dots are individual members, contours are their density, and black dots are the observed IBTrACS positions along the track (grey line). IFS-ENS's members separate into distinct clusters at intermediate leads, which an ensemble mean cannot represent.}
    \label{fig:bualoi}
\end{figure}

\section{Reanalysis setting: evaluation of cyclone forecasts on ERA5-trained models}
\label{sec:evaluation}

\subsection{Predicting cyclone locations}

Recent evaluations report cyclone position errors using the ensemble-mean track \citep{price_gencast_2024, alet_skillful_2025}. This approach quantifies the accuracy of the mean predicted position, but cannot distinguish ensembles with the same mean and different spreads or distinct possible paths. For Typhoon Bualoi (2025), IFS-ENS members separate into several clusters at intermediate lead times, which the mean alone cannot represent (Figure~\ref{fig:bualoi}). We therefore evaluate the spatial distribution of predicted locations by scoring the density assigned to the observed storm position.

We use the \textbf{TempestExtremes} tracking algorithm to detect cyclone centers in atmospheric fields and chain them into trajectories \citep{ullrich_tempestextremes_2017, ullrich2021tempestextremes} for each forecast member. Detected tracks are matched to observed tropical cyclones by comparing their times and locations. At each lead time $t$, we estimate a spatial probability density $\hat{p}_t$ from the predicted positions $\vx_{t,m}$ of that storm using kernel density estimation (KDE). We score the observed IBTrACS positions by their negative log-likelihood (NLL):
\begin{equation}
    \hat{p}_t(\vx) = \frac{1}{M_t} \sum_{m=1}^{M_t} K_h\big(\vx - \vx_{t,m}\big),
    \qquad
    \mathrm{NLL} = -\frac{1}{|\mathcal{T}|} \sum_{t \in \mathcal{T}} \log \hat{p}_t(\vy_t),
    \label{eq:location_nll}
\end{equation}
Here, $M_t$ is the number of ensemble members with an identified position for the observed storm at lead time $t$, $K_h$ is a Gaussian kernel, and $\vy_t$ is the observed position. Lower NLL means a greater probability density is assigned to that position. We score all three models at the same eligible lead times $\mathcal{T}$ and average first within each storm, then across storms, giving each storm equal weight.
NLL evaluates predicted locations only in members where TempestExtremes detects a cyclone and associates it with the observed storm. We require at least 10 such members per model at 5 or more common lead times, which leads to retaining 106 of the 125 storms. Appendix~\ref{app:reproducibility} gives the tracking, density estimation, and filtering rules.

GenCast has the lowest mean NLL and predicts a greater density of cyclone positions around the observed locations than IFS-ENS for 68\% of retained storms, and outperforms FCN3 on 62\% of storms (Table~\ref{tab:traject}). FCN3 also outperforms IFS-ENS by a smaller majority, 58\%. Although the average NLL differences are modest, both GenCast and FCN3 outperform IFS-ENS on a majority of storms. No model dominates, however: GenCast ranks first in achieving the correct position for 48\% of storms, compared with 29\% for FCN3 and 23\% for IFS-ENS. Appendix~\ref{app:stability} breaks these scores down by storm category, basin, and year.

These scores only include members for which the storm is detected and identified. As a complement, we report the \textbf{track detection rate}, the percentage of members with an identified track, averaged over all 125 storms. Rates show the same model ranking as NLL but reveal no additional skill advantage. The absence of detection can stem from forecast uncertainty about cyclone formation or persistence, model errors, or tracking limitations. Correctly assessing the good calibration of occurrence probabilities would also require considering cases where no cyclone occurred, which is outside the scope of this evaluation.

\begin{table}[t]
\centering
\caption{\textbf{Trajectory NLL and track detection rates (2021--2023).} Left: mean $\pm$ std across storms; only the lowest forecast NLL is bold. The track detection rate is the percentage of members with a detected track matched to IBTrACS, computed on all 125 storms. ERA5 is a reference-field baseline. Right: share of the 106 retained storms on which each model has the lower conditional NLL, for each pair and for all three at once.}
\label{tab:traject}
{\small\setlength{\tabcolsep}{4pt}\renewcommand{\arraystretch}{1.7}%
\begin{tabular}{@{}lcc@{}}
\toprule
\textbf{Model} & \shortstack{\textbf{Track detection}\\\textbf{rate}} & \shortstack{\textbf{Cyclone}\\\textbf{track NLL} ($\downarrow$)} \\ \midrule
GenCast & $74.9 \pm 30.9$\% & $\mathbf{13.31 \pm 1.44}$ \\
FCN3 & $73.5 \pm 31.4$\% & $13.39 \pm 1.27$ \\
IFS-ENS & $72.4 \pm 28.2$\% & $13.49 \pm 1.07$ \\ \midrule
\textit{ERA5 (baseline)} & $\mathit{94.4 \pm 23.0}$\% & --- \\ \bottomrule
\end{tabular}}\hspace{10pt}%
\begin{tabular}[c]{@{}c@{}}
{\small\textbf{GenCast has the lowest NLL on 48\% of storms}} \\[3pt]
\includegraphics{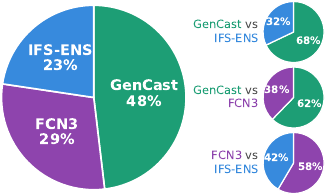}
\end{tabular}
\end{table}

We have so far assessed storm location and detection. We next examine whether the models reproduce the spatial structure and intensity of cyclone fields.

\subsection{Global scores, local skill, and intensity bias}

\paragraph{Global and cyclone-local CRPS.}
Global weather scores bear little information on tropical cyclone forecast quality as they are averaged over vast regions where no cyclone is present. We thus compute the continuous ranked probability score (CRPS) both globally and within regions enclosing the observed cyclone tracks. CRPS compares the forecast distribution of a given variable to the ERA5 value at each grid point; lower scores are better. We combine scores across seven normalized atmospheric variables. Aggregation and sampling are detailed in Appendix~\ref{app:reproducibility}.

\begin{figure}[H]
    \centering
    \includegraphics[width=0.49\linewidth]{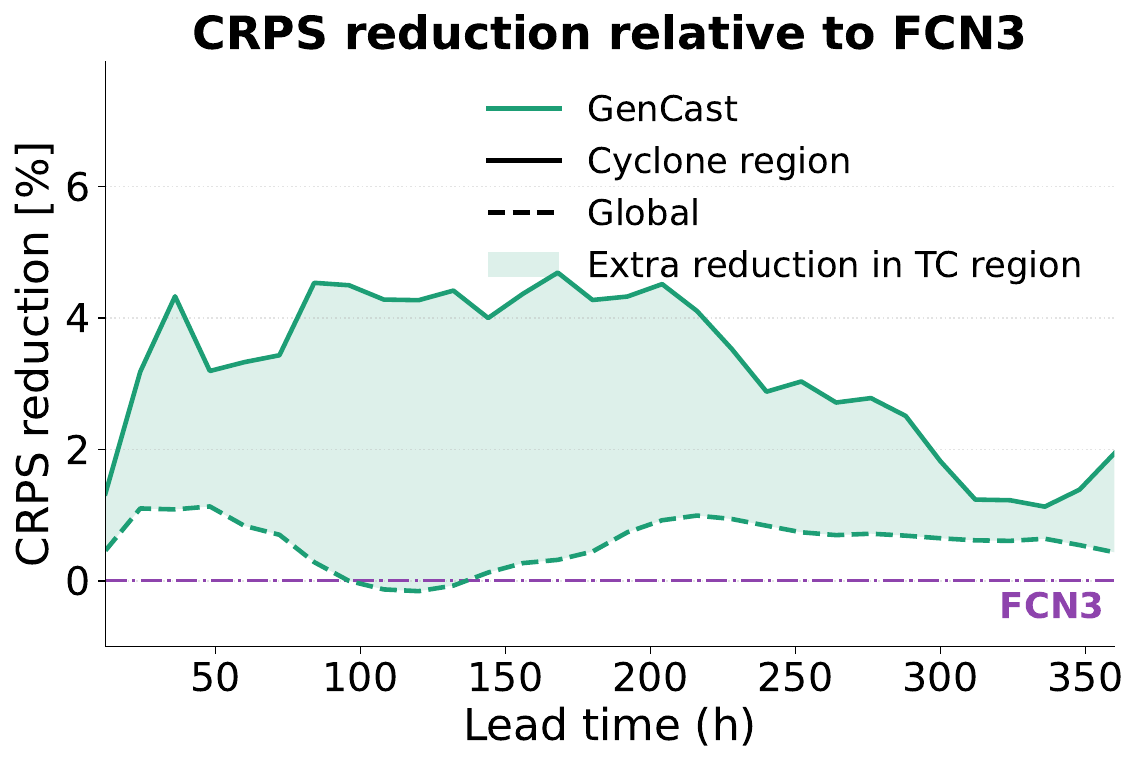}\hfill
    \includegraphics[width=0.49\linewidth]{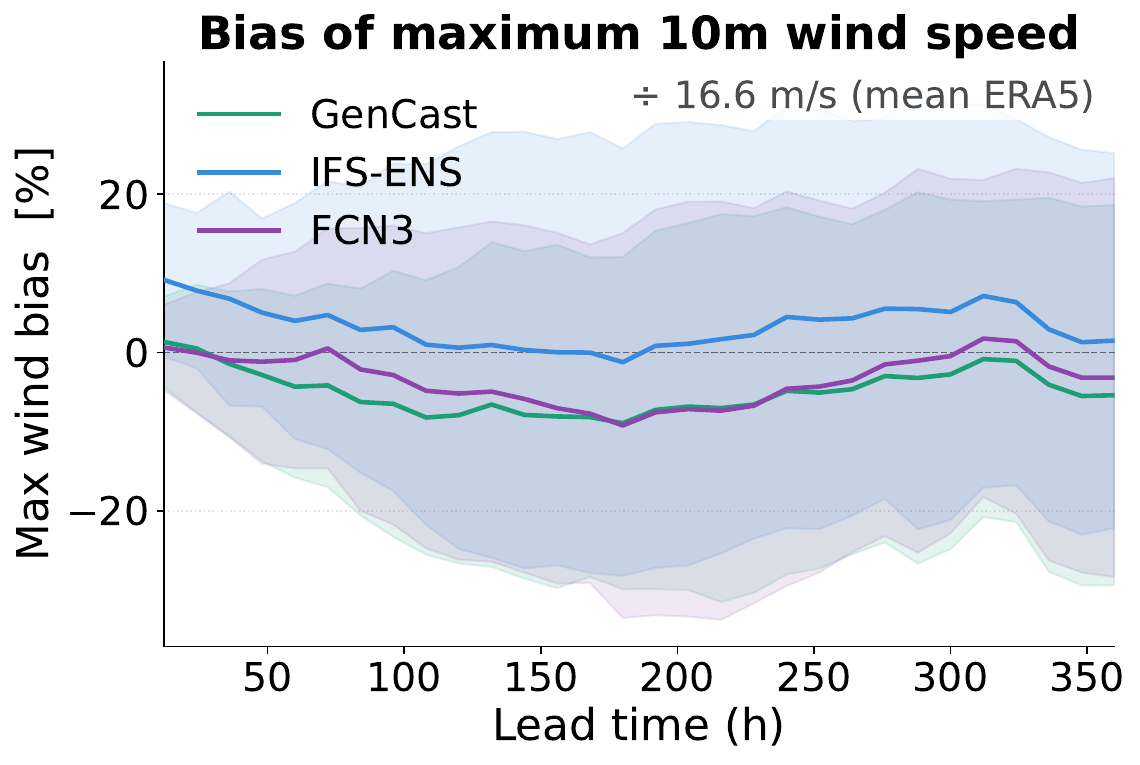}
    \caption{\textbf{Better cyclone-local CRPS need not mean less intensity bias (2021--2023, ERA5 reference).} Left: reduction of GenCast's summed normalized fair-CRPS relative to FCN3's, $100\,(1 - \mathrm{CRPS}_{\text{GenCast}}/\mathrm{CRPS}_{\text{FCN3}})$, in cyclone regions (solid) and globally (dashed); positive favors GenCast. The shaded area is GenCast's additional reduction in the cyclone region: 3--5\% from 24--220~h, versus about 1\% globally. Right: bias of the ensemble-mean maximum 10m wind speed, averaged across storms per lead and normalized by ERA5's mean maximum wind (16.6~m/s). GenCast underestimates peak winds at short and intermediate leads despite its better local CRPS.}
    \label{fig:v2_crps_bias}
\end{figure}

Globally, GenCast and FCN3 CRPS are very similar, which is consistent with \citet{bonev_fourcastnet_nodate}, even though FCN3 is specifically trained on a CRPS-based objective. Restricting the evaluation to cyclone regions shows a clear separation of the two models' performances, as GenCast achieves lower scores (Figure~\ref{fig:v2_crps_bias}, left). 

\paragraph{Peak-wind bias.}
GenCast's lower cyclone-local CRPS does not necessarily imply more accurate peak winds. To show this, we compare the ensemble mean of each member's maximum 10m wind speed with ERA5's maximum in the same region.

GenCast underestimates peak winds at short and intermediate lead times. FCN3 in contrast has a smaller mean bias up to about 150~h (Figure~\ref{fig:v2_crps_bias}, right). Beyond that, the two models exhibit similar biases. GenCast's better cyclone-local CRPS therefore coexists with a larger intensity bias. A smaller mean bias can coexist with compensating estimation errors on individual storms.

\paragraph{Interpreting the scores.}
These results show two limits to the characterization of cyclone forecasts quality on aggregate scores only. Global averages conceal differences around cyclones, while even cyclone-local scores can conceal a weakness in the representation of peak winds. GenCast's lower local CRPS supports better agreement with ERA5 in tropical cyclone regions overall, but does not establish that every property of the storm is better predicted. Nor does this comparison identify which errors account for its advantage. This motivates assessing spatial agreement separately from intensity in the operational setting.

An important note is that ERA5 itself underestimates cyclone maximum winds \citep{dulac_assessing_2024}. Agreement with ERA5 consequently does not establish realistic intensity, nor does IFS-ENS's positive bias establish that its storms are too strong. We next examine whether models fine-tuned on operational analyses improve cyclone tracking and whether their intensity biases persist against HRES-fc0, while assessing spatial agreement separately from intensity.

\section{Operational setting: tracking, intensity, and spatial agreement for models fine-tuned on operational analyses}
\label{sec:operational}

\paragraph{Cyclone tracking.}
FGN assigns a greater probability density to the observed storm locations than IFS-ENS on approximately 90\% of the 41 retained storms and achieves the lowest mean NLL (Table~\ref{tab:traject_fgn}). Operational GenCast also outperforms IFS-ENS on 78\% of storms, while FGN's advantage over GenCast is smaller, at 61\%. FGN ranks first among all three models on 59\% of storms, compared with 34\% for GenCast and 7\% for IFS-ENS. Both AI models therefore improve tracking on most evaluated storms, with FGN ahead of GenCast.

The detection rates follow the NLL ordering, with both AI models ahead of IFS-ENS. Unlike in 2021--2023, where detection rates were close, the operational comparison shows a clear separation. This reflects both higher detection for operational GenCast and lower detection for IFS-ENS on the 2025 storms. The changing numerical baseline also prevents attributing differences between settings to AI fine-tuning alone. Whether its lower detection rate reflects the storm sample, forecast uncertainty, or differences in forecast structure and tracking remains unresolved.

\begin{table}[H]
\centering
\caption{\textbf{Trajectory NLL and track detection rates (2025).} Left: mean $\pm$ std across storms; only the lowest forecast NLL is bold. FGN and GenCast have similar track detection rates, computed on all 53 storms. HRES-fc0 is a reference-field baseline. Right: share of the 41 retained storms on which each model has the lower conditional NLL, for each pair and for all three at once. FGN's NLL advantage over GenCast is smaller than its advantage over IFS-ENS.}
\label{tab:traject_fgn}
{\small\setlength{\tabcolsep}{4pt}\renewcommand{\arraystretch}{1.7}%
\begin{tabular}{@{}lcc@{}}
\toprule
\textbf{Model} & \shortstack{\textbf{Track detection}\\\textbf{rate}} & \shortstack{\textbf{Cyclone}\\\textbf{track NLL} ($\downarrow$)} \\ \midrule
FGN & $78.9 \pm 34.6$\% & $\mathbf{12.13 \pm 1.21}$ \\
GenCast Operational & $78.0 \pm 33.9$\% & $12.55 \pm 2.08$ \\
IFS-ENS & $66.2 \pm 36.4$\% & $13.06 \pm 1.57$ \\ \midrule
\textit{HRES-fc0 (baseline)} & $\mathit{96.2 \pm 19.1}$\% & --- \\ \bottomrule
\end{tabular}}\hspace{10pt}%
\begin{tabular}[c]{@{}c@{}}
{\small\textbf{FGN beats IFS-ENS on 90\% of storms}} \\[3pt]
\includegraphics{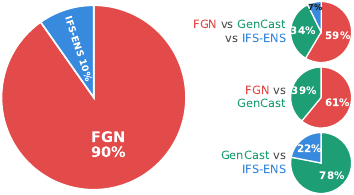}
\end{tabular}
\end{table}

\begin{figure}[t]
    \centering
    \includegraphics[width=0.49\linewidth]{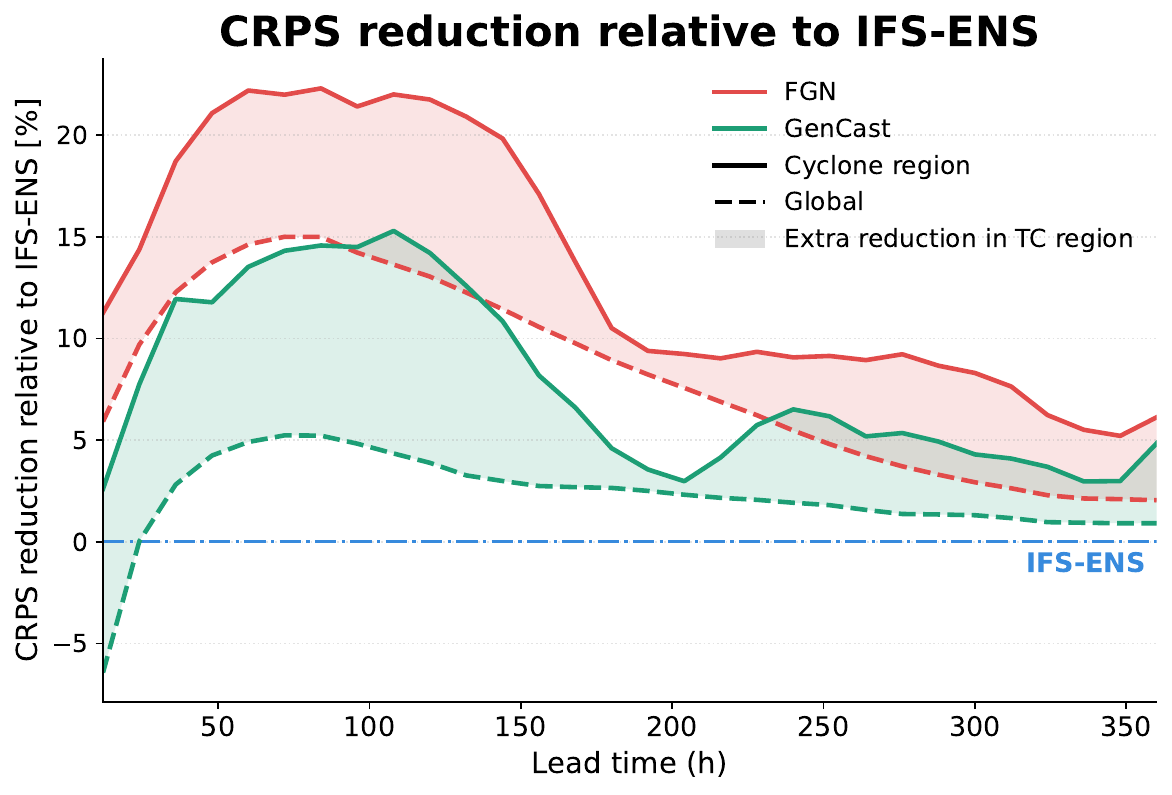}\hfill
    \includegraphics[width=0.49\linewidth]{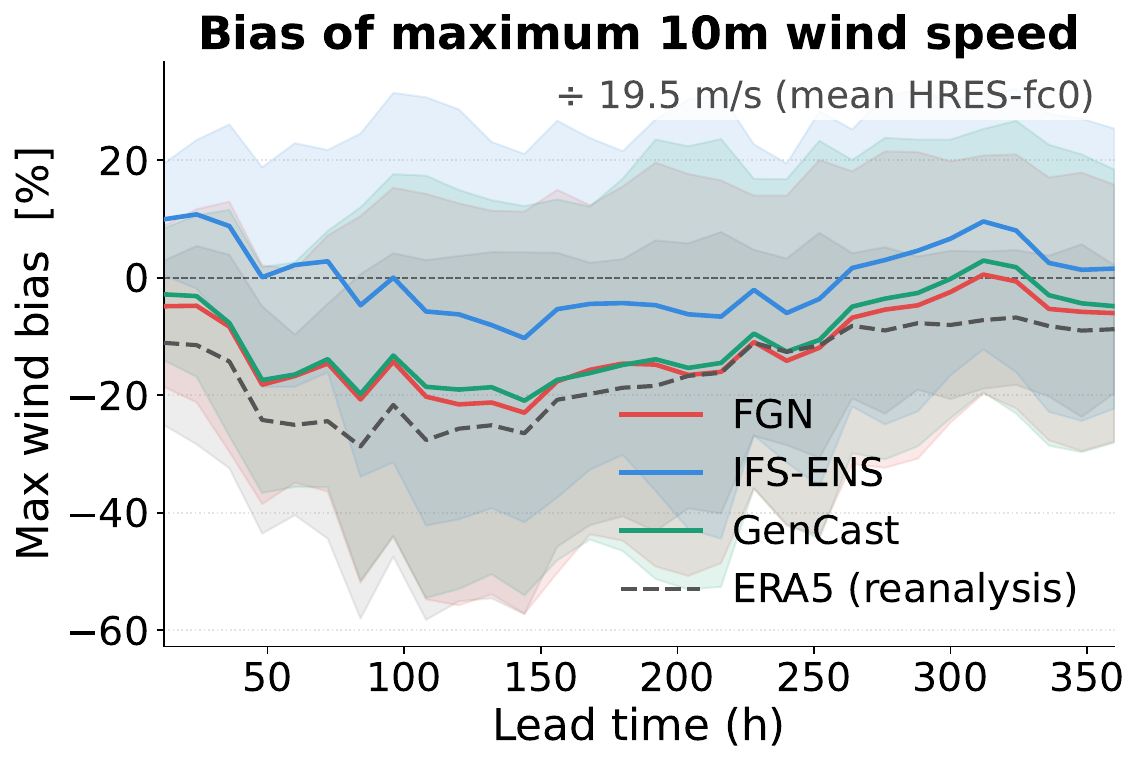}
    \caption{\textbf{Better operational CRPS coexists with persistent intensity bias (2025, HRES-fc0 reference).} Left: fair-CRPS reduction of each AI model relative to IFS-ENS, $100\,(1 - \mathrm{CRPS}_{\text{model}}/\mathrm{CRPS}_{\text{IFS-ENS}})$, in the cyclone region (solid) and globally (dashed); positive favors the AI model. The shaded area is a model's additional reduction in the cyclone region. FGN leads GenCast in both domains, and both improve more on IFS-ENS in the cyclone region than globally, though GenCast's global CRPS is above IFS-ENS's at the first lead. Right: bias of the ensemble-mean maximum 10m wind speed, averaged across storms per lead and normalized by HRES-fc0's mean maximum wind (19.5~m/s), with ERA5 for comparison. FGN and GenCast have similar negative peak-wind biases at short and intermediate leads despite their different CRPS.}
    \label{fig:v2_operational_intensity}
    \label{fig:v2_operational_crps}
\end{figure}

\paragraph{Global and cyclone-local CRPS.}
The advantage extends beyond storm-center locations. FGN has a lower CRPS than both GenCast and IFS-ENS, globally and around cyclones. Both AI models show a stronger improvement compared to IFS-ENS within cyclone regions than globally (Figure~\ref{fig:v2_operational_intensity}, left). Given concerns about learned models at extremes, the larger relative gains around cyclones are encouraging. They establish forecast skill without identifying the physical mechanisms learned. Appendix~\ref{app:spectra} compares how the models' energy is distributed across spatial frequencies.

The field scores still combine errors in intensity and spatial placement. We next examine whether the models locate the strongest winds and lowest pressures more accurately.

\paragraph{Spatial agreement.}
A forecast may correctly locate the strongest winds while underestimating their intensity. To evaluate placement independently of magnitude, we compare the regions of strongest winds and lowest pressures across forecasts against those in HRES-fc0. These regions are defined using percentile thresholds computed from the pooled values of all ensemble members for a given model, ensuring that a systematic intensity bias does not influence the comparison. The resulting masks therefore capture the relative placement and shape of extreme features, rather than areas exceeding a fixed hazardous wind speed threshold. We then evaluate these masks using ensemble probabilities via the Brier score, and spatial agreement via overlap and distance metrics; formal definitions are provided in Appendix~\ref{app:reproducibility}.

\begin{figure}[t]
    \centering
    \includegraphics[width=\linewidth]{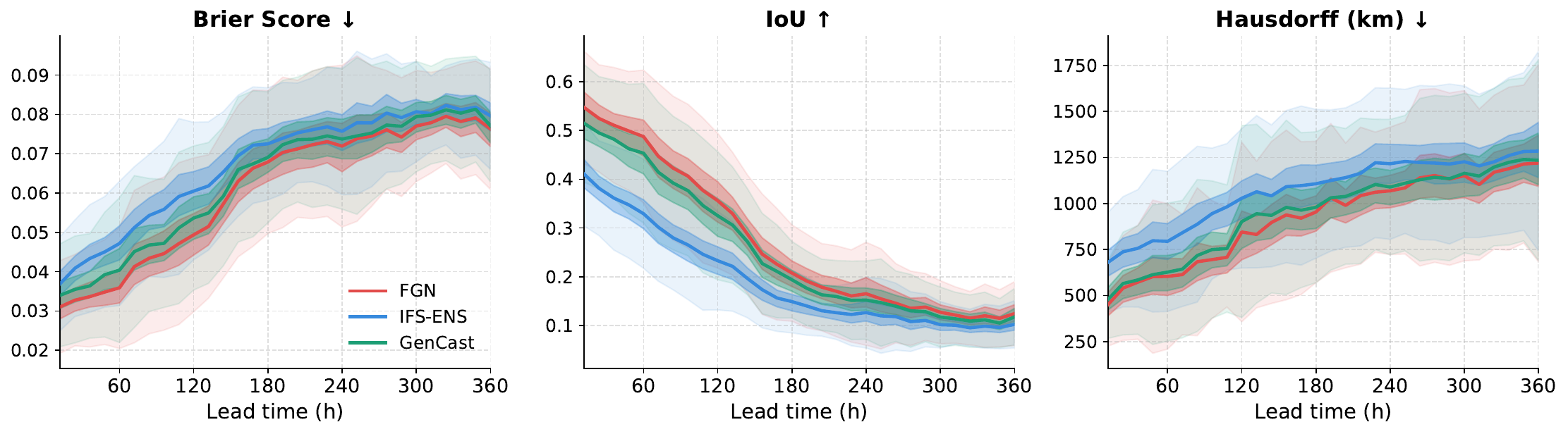}
    \caption{\textbf{Location metrics by lead time (2025, 10m wind, HRES-fc0 reference).} Brier score (left), IoU (middle), and Hausdorff distance in kilometres (right), at the 90th percentile. All members of a model share one threshold, the percentile of the pooled member values; the reference uses its own.}
    \label{fig:v2_location_lead}
\end{figure}

FGN performs best on the spatial diagnostics, followed by GenCast and IFS-ENS, at nearly all lead times (Figure~\ref{fig:v2_location_lead}). Comparisons of storm footprints, the strongest winds and lowest pressures reached at each location over the forecast window, give the same ordering across all reported metrics for both variables (Appendix~\ref{app:operational_diagnostics}, Figure~\ref{fig:location-footprint}). The tracking advantage thus extends to the placement of strong winds and low pressures, even after removing sensitivity to a uniform intensity bias.

\paragraph{Intensity and calibration.}
Better placement does not bring a comparable improvement in peak winds. Using the same intensity diagnostic as in Section~\ref{sec:evaluation}, FGN and GenCast have similar negative wind-intensity biases at short and intermediate leads, despite FGN's lower CRPS and better spatial scores (Figure~\ref{fig:v2_operational_intensity}, right). Both models tend to underestimate maximum 10m winds relative to HRES-fc0, whereas IFS-ENS remains closer to the reference. Across members, track and intensity errors are positively correlated, though only moderately so (Appendix~\ref{app:track_intensity}).

ERA5 has a larger negative bias compared to HRES-fc0 than either AI model, while IFS-ENS bias remains below 10\% in absolute value. Operational fine-tuning has not eliminated the AI models' weak-wind bias. Its reduction towards later forecast lead times may partly reflect storm dissipation rather than improved forecasts of active storms. Rank histograms also show calibration errors (Appendix~\ref{app:rank_histograms}). The operational models thus improve the placement of strong conditions while retaining intensity and calibration errors.

Despite these magnitude biases, we next examine whether the predicted winds and pressures co-vary consistently with the reference data, by analysing their joint distributions.

\section{Do the models reproduce the wind--pressure relationship?}
\label{sec:storm_structure}

Stronger tropical cyclones are generally characterised by higher wind speeds and lower central pressure. We therefore examine whether the models reproduce this wind-pressure relationship, even when the predicted storm intensity is systematically underestimated. This is particularly relevant to FGN: its CRPS  training objective evaluates each variable and location separately, without directly enforcing their joint dependence. GenCast, by contrast, relies on diffusion to learn the joint distribution of atmospheric fields \citep{alet_skillful_2025, price_gencast_2024}.

\begin{figure}[t]
    \centering
    \begin{subfigure}[t]{0.49\linewidth}
        \centering
        \includegraphics[width=\linewidth]{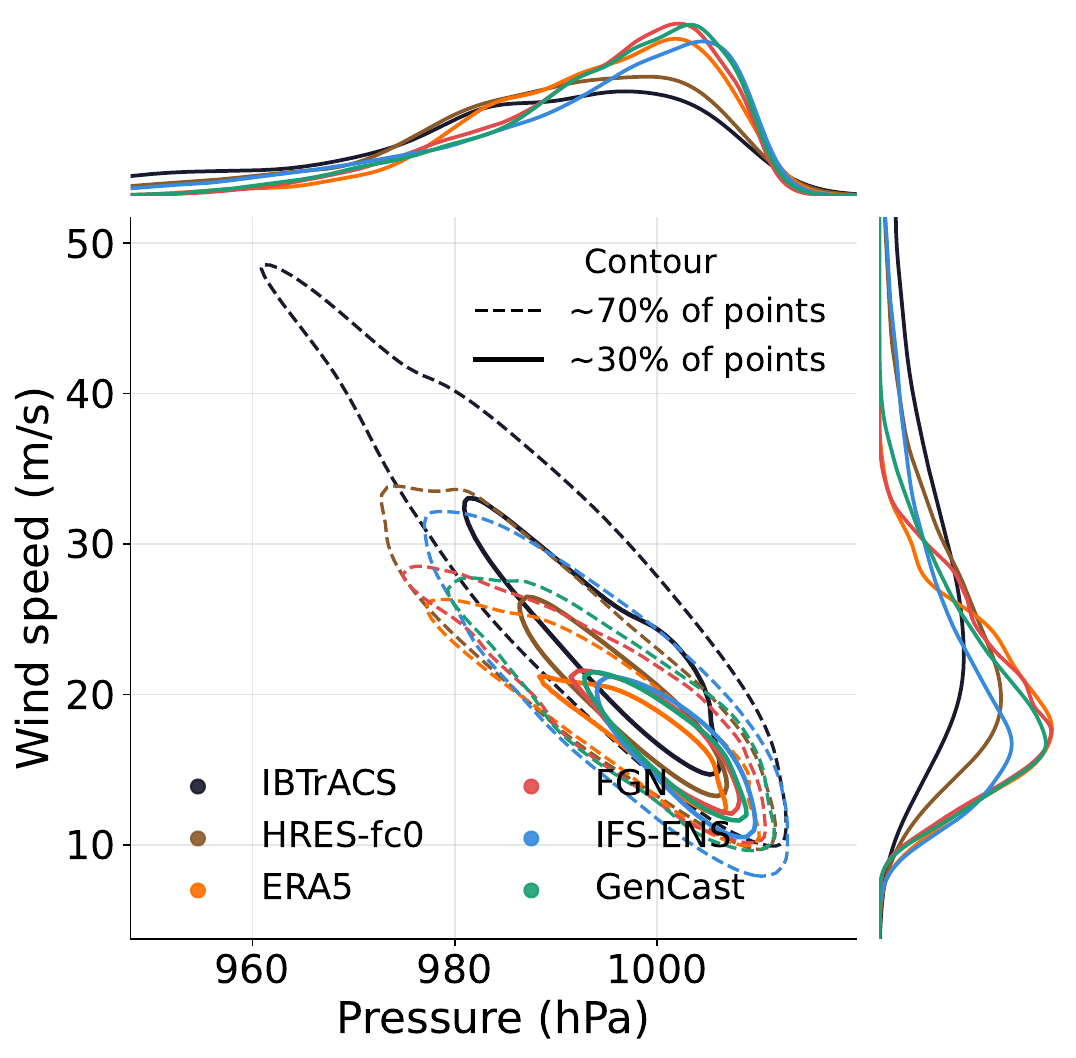}
        \caption{Joint wind--pressure distributions and their marginals.}
    \end{subfigure}\hfill
    \begin{subfigure}[t]{0.49\linewidth}
        \centering
        \includegraphics[width=\linewidth]{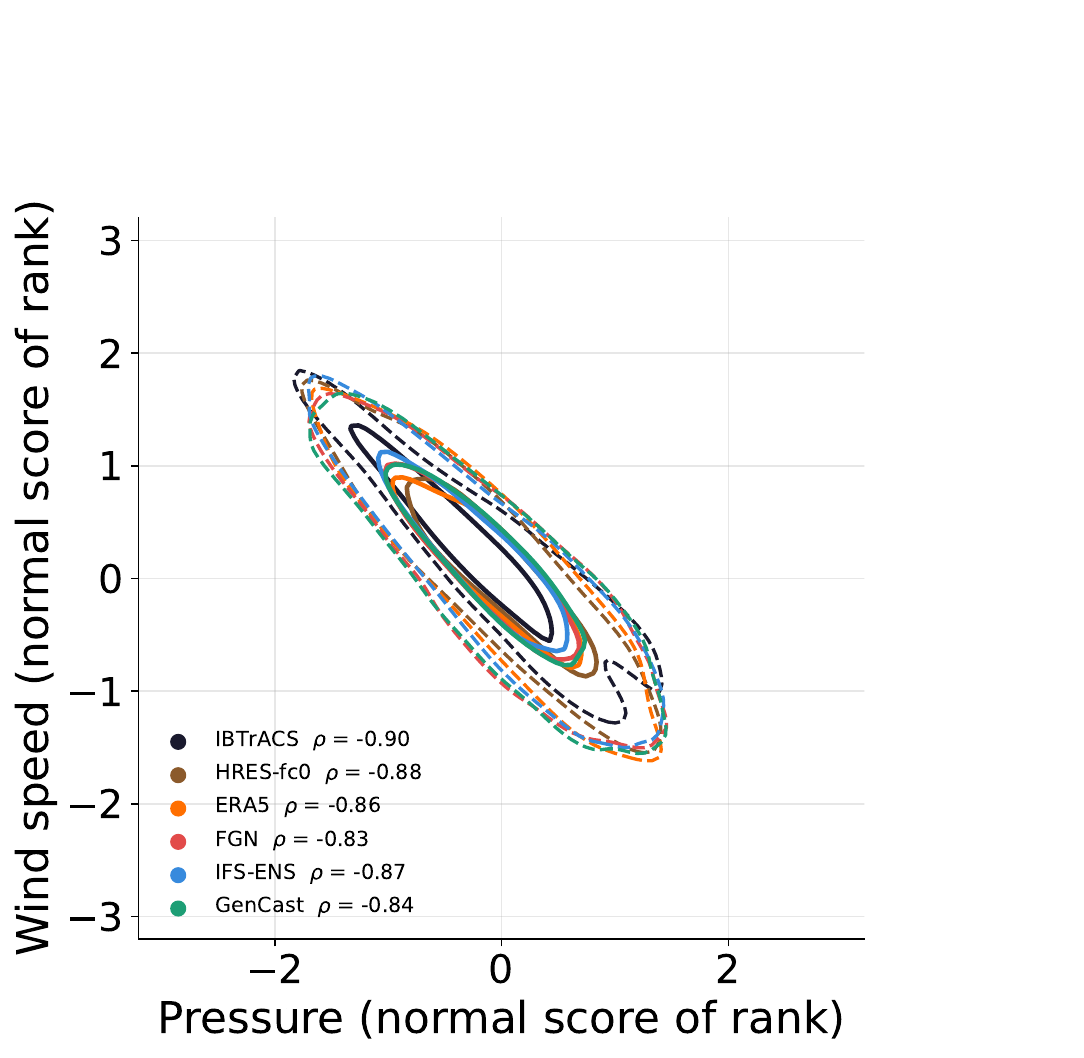}
        \caption{Rank dependence (copula), marginal biases removed; Spearman's $\rho$ in the legend.}
    \end{subfigure}
    \caption{\textbf{Operational AI forecasts differ from HRES-fc0 in storm intensity, not in how wind follows pressure (2025).} Each source contributes minimum pressure and maximum wind speed at timesteps matched to IBTrACS observations, pooled over all lead times. (a) Contours of the pooled joint distributions, with marginal densities. (b) The same points ranked within each source and converted to standard normal scores, so all marginals are identical and the contours show only the dependence. FGN and GenCast have weaker winds and shallower pressures than their fine-tuning reference, HRES-fc0, with marginals close to ERA5's (a), but every gridded source has nearly the same dependence (b). Dashed and solid contours enclose about 70\% and 30\% of each source's points; IBTrACS's sustained winds differ in definition from the others' gridded instantaneous winds.}
    \label{fig:v2_joint_operational}
\end{figure}

\paragraph{Comparing predicted storms.}
At each forecast time, we take the strongest 10m wind and lowest sea-level pressure within the cyclone region, giving one wind--pressure pair per ensemble member. These extrema depend on how the predicted fields vary across space; their pairing also tests how wind and pressure vary together. We compare their distributions to ERA5, HRES-fc0, and IBTrACS at times with valid observed wind and pressure. The main figure pools storms, members, and lead times; Appendix~\ref{app:reproducibility} gives the sampling details. FGN and GenCast produce weaker winds and shallower pressure minima than HRES-fc0, with distributions closer to ERA5 despite operational fine-tuning (Figure~\ref{fig:v2_joint_operational}a). We therefore assess whether these models nonetheless preserve the expected physical relationship between wind intensity and central pressure.

\paragraph{Separating dependence from intensity.}
To isolate the dependence structure from marginal biases, we rank wind and pressure separately within each dataset and transform the ranks to standard normal scores. This rank-based transformation renders the marginal distributions identical across datasets, while preserving their joint dependence structure, represented by the copula. The copulas are found to be consistent across models and references (Figure~\ref{fig:v2_joint_operational}b): FGN and GenCast have rank correlations of $-0.83$ and $-0.84$, compared with $-0.88$ for HRES-fc0. The qualitative result also holds in the earlier evaluation (Appendix~\ref{app:wind_pressure}) and across separate lead-time ranges. IBTrACS supplies observational context, although its sustained winds differ from gridded instantaneous winds.

FGN therefore reproduces a wind--pressure dependence structure comparable to that of GenCast, despite being trained with CRPS applied separately to each variable and grid point, which constrains their conditional marginal distributions but not their joint dependence. The coexistence of negatively biased wind speeds and pressure minima with a realistic wind–pressure association suggests that intensity errors and dependence errors are distinct. While this does not imply that the models have learned an explicit physical mechanism, nor that wind speed is correctly predicted at a given pressure level, it allows intensity biases to be clearly separated from errors in the dependence structure.

\section{Related work}
\label{sec:related_work}

\paragraph{AI forecasts of extreme weather.}
Evaluations are mixed: \citet{olivetti_data-driven_2024} find competitive skill across extreme events, varying by event, region, and lead time; \citet{zhang_numerical_2025} find numerical forecasts superior on record-breaking extremes. For cyclones, \citet{demaria_evaluation_2024} show competitive tracking alongside substantial intensity underestimation. We revisit this distinction for recent probabilistic models, asking whether similar global scores conceal differences in cyclone fields.

\paragraph{Probabilistic cyclone evaluation.}
TCBench \citep{gomez_tcbench_2026}, the closest benchmark, evaluates one-to-five-day track and intensity forecasts, including probabilistic scores, rapid intensification, and postprocessing. GEFS \citep{zhou_development_2022} outperforms its evaluated AI ensembles on track CRPS: probabilistic evaluation does not uniformly favor AI. Different models, seasons, references, and scoring protocols preclude direct comparison with our rankings. We complement this work with location-density scores, track detection rates, and surrounding-field diagnostics. CRPS scores forecast distributions without separating intensity from spatial errors; our intensity and quantile-footprint analyses examine these separately without additively decomposing CRPS.

\paragraph{Physical relationships in learned forecasts.}
Recent work studying cyclone dynamics and thermodynamics has examined AI-based weather models through the lens of cyclone physics: \citet{sahu_evaluating_2025} find realistic spatial patterns but weak intensity while \citet{bano-medina_are_2025} assess the sensitivity of learned and numerical models to initial conditions for extratropical cyclone Xynthia. Our joint wind--pressure analysis complements these studies by providing a distributional diagnostic aggregated across storms and ensemble members, using reanalysis, operational analyses, and observations to identify which physical relationships are preserved and which biases persist. However, agreement with a known physical relationship does not establish the underlying learned mechanism, just as resemblance to pretraining data does not imply a causal effect of the training data.

\section{Conclusion}
\label{sec:conclusion}

Our benchmark shows that recent probabilistic models can outperform numerical forecasts on cyclone location and surrounding fields, while still exhibiting intensity biases and calibration errors. The presence of systematically weak storms alone is therefore insufficient to conclude a general failure in cyclone representation. FGN's wind--pressure dependence structure, comparable to that of GenCast, despite being trained solely on conditional marginals, further shows that meaningful joint structure can emerge implicitly, without being explicitly enforced by the training objective.

These findings suggest more targeted routes to improvement. Comparing model outputs against observations and higher-resolution analyses could help disentangle errors inherited from training targets from limitations of the models themselves. Potential remedies such as intensity correction, loss functions that emphasize extreme winds, and improved uncertainty quantification should be evaluated not only for their effect on intensity, but also for whether they preserve spatial skill and wind–pressure dependence. Finally, comparing marginal and joint training objectives under matched architectures and compute budgets could clarify the extent to which the recovered dependence structure arises from the training objective rather than from the model's inherent structural constraints.

Progress on extreme weather forecasting with AI models should consequently be assessed beyond aggregate global scores. Evaluation across additional seasons, including periods without observed cyclones, would test the robustness of these gains and assess the calibration of cyclone-occurrence probabilities. Decomposing forecast performance into location, intensity, uncertainty, and dependence makes such evaluations actionable for model development: it identifies what needs improvement and what should be preserved, while providing more specific evidence about what learned forecasts capture of extreme weather.

\subsection*{AI use statement}
We used generative AI tools, including a large language model coding assistant, to revise plotting code written by the authors, to make exploratory plots, to clean and comment our code for a reusable public release, to draft and edit parts of the text, and for grammar correction. We did not use generative AI to create data or to design the study and its conceptual framework, and the interpretation of the results was done by the authors. We reviewed all AI-assisted work and take full responsibility for the final content of this paper, including text, claims, code, and figures produced with the aid of generative AI.

\subsection*{Ethics statement}
This work evaluates publicly released models and archived forecasts on past storms; it involves no human subjects or personal data. All forecasts, reanalyses, and observations are used under the licenses listed in Appendix~\ref{app:reproducibility}. Tropical cyclone forecasts inform evacuation and preparation decisions, so our results should not be read as guidance on using any of these models for operational warnings: they are a retrospective evaluation of selected storms, and operational use would require evaluation under operational conditions, including cases without cyclones. Generating the forecasts we ran ourselves required more than 6,000 H100 GPU-hours.

\clearpage
\appendix

\section{Evaluation protocol and reproducibility}
\label{app:reproducibility}

\paragraph{Events, initialization, and ensemble size.}
Selection uses each storm's maximum IBTrACS \texttt{USA\_WIND}, with a threshold of at least 64 knots, i.e.\ category 1 to 5 on the Saffir--Simpson scale, yielding 125 storms in 2021--2023 and 53 in 2025. A storm's first time is its first IBTrACS record with a reported \texttt{USA\_WIND}, and each forecast starts 12 hours before it, after moving any 06:00/18:00 UTC start to the preceding 00:00/12:00 cycle, since IFS-ENS forecasts initialized at 06:00/18:00 UTC do not cover the full 15-day window in the sources we used. Selection by lifetime maximum does not mean the storm is already at hurricane strength at initialization; this protocol evaluates selected known events, not unconditional cyclone occurrence. All forecasts span 15 days. We retain 50 members per model in 2021--2023. Missing individual files in the 2025 GenCast and IFS-ENS archives motivate a uniform 49-member comparison for all three models. IFS-ENS uses perturbed members, excluding the control. Equal membership is not an equal-compute comparison.

\paragraph{Forecast and reference sources.}
We ran GenCast and FCN3 ourselves and retrieved the other forecasts from public archives:
\begin{itemize}
    \item \textbf{IFS-ENS}: the 50 perturbed members, excluding the control; for 2021--2023 from TIGGE
    \citep{bougeault2010thorpex} (ECMWF, CC BY 4.0), already on the $0.25^\circ$ grid, and for 2025 from
    ECMWF Open Data (CC BY 4.0) through Earth2Studio's \texttt{IFS\_ENS\_FX} interface. The IFS cycles in operation are listed in Table~\ref{tab:ifs-cycles}.
    \item \textbf{GenCast}: DeepMind's implementation (Apache 2.0) with the released $0.25^\circ$, 13-level checkpoint
    trained on ERA5 1979--2018 (CC BY-NC-SA 4.0), initialized from ERA5 (see Reference fields below); about 30 minutes
    per 15-day member on H100/H200 GPUs, which limited the number of storms.
    \item \textbf{FCN3}: NVIDIA's released weights (Hugging Face \texttt{nvidia/fourcastnet3}, revision \texttt{76ef0c6}; Apache 2.0)
    run through Earth2Studio (Apache 2.0),
    initialized from ARCO-ERA5 \citep{carver2023arco}; about one minute per member on H200 GPUs.
    \item \textbf{FGN}: precomputed forecasts from the WeatherNext~2 archive,
    \path{weathernext/weathernext_2_0_0/zarr/2025_to_present} (CC BY 4.0).
    \item \textbf{GenCast operational}: precomputed forecasts from the WeatherNext Gen archive,
    \path{weathernext/126478713_1_0/zarr/126478713_2024_to_present} (CC BY 4.0).
\end{itemize}

\paragraph{Model size and compute.}
The models also differ in size and in the compute required for inference. GenCast has a single network of about 57M parameters, each FGN network has 180M parameters, and FCN3 has 711M parameters \citep{alet_skillful_2025, bonev_fourcastnet_nodate}. GenCast requires significantly more compute at inference (20 diffusion solver steps, 39 network evaluations) for a 12-hour forecast step \citep{price_gencast_2024}, whereas FGN and FCN3 generate a 6-hour step in a single forward pass. However, GenCast reportedly has a lower training budget than FGN \citep{alet_skillful_2025, price_gencast_2024}.

IBTrACS (NOAA NCEI, public domain) supplies observed positions in both settings; ERA5 and HRES-fc0 supply the gridded references.
These are different verification targets, so absolute field scores should not be compared across cohorts.

\paragraph{Reference fields.}
ERA5 (Copernicus Climate Change Service, CC BY 4.0) for 2021 was retrieved from WeatherBench~2 \citep{rasp_weatherbench_2024}, and for 2022--2023 from the
Copernicus Climate Data Store, at $0.25^\circ$. HRES-fc0 was retrieved from
ECMWF Open Data (CC BY 4.0) through Earth2Studio's \texttt{IFS} interface: for each valid time, the 0-hour state of the
operational high-resolution forecast initialized at that time, at $0.25^\circ$, i.e.\ after data assimilation of observations from $\pm$3 hours.

\begin{table}[htbp]
\centering
\caption{\textbf{IFS-ENS cycles in operation over the evaluation periods}, from ECMWF's record of changes
to the forecasting system. The 2021--2023 storms span four cycles: IFS-ENS is not a fixed baseline within
that period, and forecasts after the 48r1 upgrade of June 2023 come from a model with half the grid spacing.
All 2025 forecasts come from 49r1.}
\label{tab:ifs-cycles}
\small
\begin{tabular}{@{}llll@{}}
\toprule
\textbf{Cycle} & \textbf{Implemented} & \textbf{ENS grid spacing} & \textbf{Levels} \\ \midrule
47r1 & 30 June 2020 & $\sim$18~km & 91 \\
47r2 & 11 May 2021 & $\sim$18~km & 91 $\rightarrow$ 137 \\
47r3 & 12 October 2021 & $\sim$18~km & 137 \\
48r1 & 27 June 2023 & 18 $\rightarrow$ 9~km & 137 \\
49r1 & 12 November 2024 & 9~km & 137 \\
\bottomrule
\end{tabular}
\end{table}

\paragraph{Common fields and evaluation window.}
All field comparisons use a $721\times1440$ latitude--longitude grid at $0.25^\circ$ spacing and 30 forecast times, 12--360 hours at 12-hour intervals. The 12-hour step is that of the coarsest model, GenCast; the models with 6-hour output are subsampled to it so that all models are compared at the same times. The seven CRPS variables are mean sea-level pressure, the two 10m wind components, geopotential at 300 and 500 hPa, and the two 850 hPa wind components. Wind speed for the intensity and spatial diagnostics is computed as $\sqrt{u_{10}^2+v_{10}^2}$. Local scores use a fixed bounding box enclosing the observed track with a $2.25^\circ$ margin. This region is shared across models; it is not centered on each predicted track. Field scores use available fields irrespective of whether the tracker detects a cyclone. Footprints reduce the stored forecast window in time by a maximum for wind or a minimum for pressure; the field pipeline does not apply the active-storm time mask used for the joint-distribution analysis below.

\paragraph{Track extraction and matching.}
Tracks are extracted with TempestExtremes \citep{ullrich_tempestextremes_2017, ullrich2021tempestextremes},
applied to the 12-hourly fields, in two steps. \texttt{DetectNodes} finds candidate storm centres at each time:
minima of mean sea-level pressure enclosed by a closed pressure contour and a warm-core contour. We use the
settings recommended for tropical cyclones in ERA5 \citep{ullrich2021tempestextremes}:
\begin{quote}\small\ttfamily
searchbymin: msl; mergedist: 6.0\\
closedcontourcmd: msl,200.0,5.5,0;\\
\hspace*{1em}\_DIFF(z300,z500),-58.8,6.5,1.0\\
outputcmd: msl,min,0;\_VECMAG(u10m,v10m),max,2.0;zs,min,0
\end{quote}
\texttt{StitchNodes} then joins candidates at consecutive times into tracks. Its recommended settings assume
6-hour steps; we adapt them to 12 hours as \citet{price_gencast_2024} do for GenCast:
\begin{quote}\small\ttfamily
range: 12.0; mintime: 54h; maxgap: 24h\\
threshold: wind,>=,10.0,5;lat,<=,50.0,5;\\
\hspace*{1em}lat,>=,-50.0,5;zs,<=,150.0,5
\end{quote}
\texttt{range} is the largest distance, in degrees, a storm centre may move between consecutive steps and stay
in the same track; it is raised from 8$^\circ$ to 12$^\circ$, which \citet{price_gencast_2024} chose to balance
cutting off fast-moving storms against joining separate storms. The last number of each \texttt{threshold} clause counts time steps, so it is halved from
10 to 5 to keep the same 60-hour duration; \texttt{mintime} and \texttt{maxgap} are in hours and unchanged.

Each extracted trajectory is matched to IBTrACS using its first four positions: all four timestamps must be
present in the candidate IBTrACS track, and the mean great-circle distance must be below 500~km; the closest
qualifying track is selected. Matching on the start of the track tests whether the model forecasts the right
storm, not how far it later drifts, which the metrics measure. Members with no matched trajectory do not enter the location density. The track detection rate of storm $s$ is
\begin{equation}
R_s=\frac{\#\{\text{members with a track matched to } s\}}{N},
\end{equation}
and the tables report the mean and standard deviation of $R_s$ across all storms of the cohort (125 or 53), not only those retained for the NLL. For the reference rows, the tracker is applied to the ERA5 or HRES-fc0 fields themselves, so $R_s$ is 1 if a matched track is found and 0 otherwise.

\paragraph{Location density.}
The NLL scores how much probability density each ensemble places on the observed storm position. For each storm and lead time $t$, the positions $\vx_{t,m}$ of the $M_t$ members with a matched track and the IBTrACS position $\vy_t$ are mapped to a local azimuthal-equidistant projection in kilometers, centered on the median of the storm's observed and predicted positions, so that distances are comparable at all latitudes. The storm's NLL is then Equation~\ref{eq:location_nll}: $\hat p_t$ is an isotropic Gaussian KDE of the member positions, logarithms are natural, and $\mathcal{T}$ is the set of retained lead times (below).

\paragraph{KDE bandwidth.}
The bandwidth is chosen for each storm, model and lead time by five-fold cross-validation over 100 values, from 13.9~km (half a grid cell) up to five times the ensemble spread, or 55.6~km if that is larger; the spread is the standard deviation of the member positions, averaged over the two axes. The lower bound matters because TempestExtremes places storm centres on grid points: a smaller bandwidth would put a separate peak on each grid point instead of describing the ensemble spread.

\paragraph{Retained storms and averaging.}
A lead time enters $\mathcal{T}$ only if all three models have at least ten matched members at that time, and a storm is retained if it has at least five such times: 106 storms in 2021--2023 and 41 in 2025. All models are therefore scored on the same storms and times. Tables report the mean and standard deviation of the storm NLLs across retained storms, and pairwise win rates compare them storm by storm.

\paragraph{Fair-CRPS and normalization.}
At a grid point with reference $y$ and ensemble $x_1,\ldots,x_N$, we use
\begin{equation}
\operatorname{fCRPS}(x,y)=\frac{1}{N}\sum_i|x_i-y|
-\frac{1}{2N(N-1)}\sum_{i\ne j}|x_i-x_j|.
\end{equation}
The usual ensemble estimator divides the second term by $2N^2$ and depends on ensemble size; the fair form estimates the score of the distribution the members are drawn from \citep{zamo_estimation_2017}.
Each variable is divided by its fixed ERA5 1979--2018 standard deviation, using the same normalization file in both cohorts. Scores are area-weighted within the evaluation domain and summed across the seven variables; this is the sum of marginal scores, not a joint multivariate score. At each lead, the plotted curves average across storms with all three models available. Model ratios divide these mean curves, rather than averaging storm-specific ratios. The field sample is not restricted to the storms retained for NLL.

\paragraph{Intensity bias and calibration.}
For storm $s$ and lead $t$, let $v_{s,t,m}$ be member $m$'s maximum wind within the local region and $v^*_{s,t}$ the reference maximum. We average $b_{s,t}=N^{-1}\sum_m v_{s,t,m}-v^*_{s,t}$ across storms at each lead. The plotted percentage is $100\,\overline b_t/\overline v^*$, where the denominator is the reference maximum averaged over the available storms and leads in that cohort (16.6 m/s for ERA5 and 19.5 m/s for HRES-fc0, rounded). It is not an average of storm-specific percentage errors. Shading gives one cross-storm standard deviation under the same scaling. Rank histograms count ensemble values strictly below the reference at each grid point, area-weight the rank frequencies, and average across storms and the displayed lead-time groups; ties are not randomized.

\paragraph{Spatial masks.}
The reference is thresholded at its own percentile; all members of a model share one threshold, the same percentile of their pooled values, computed separately for each storm, variable, and footprint or lead time. This removes the model's uniform intensity bias while keeping intensity differences between its members. A wind threshold at percentile $\alpha$ retains values above that percentile; pressure uses values below percentile $100-\alpha$. Footprint percentiles range from 50 to 99; the fields at each lead time use the 90th percentile. Ensemble mask frequencies enter area-weighted Brier scores, with a separate score restricted to reference-positive pixels. Geometric scores compare each member's mask with the reference (error) and every pair of members (spread), and are averaged over members or pairs, then storms. Hausdorff distance and G-score use kilometre distances from a local equirectangular projection of each storm's region, so equal displacements count equally at all latitudes. Quantile masks remove uniform intensity shifts but do not separate spatial displacement from shape differences.

\paragraph{Spatial metrics.}
Let $T$ be the reference mask, $M_1,\ldots,M_N$ the member masks, $P_k=N^{-1}\sum_i M_{i,k}$ the fraction of members active at grid cell $k$, and $w_k$ the cell area. The Brier score and its restriction to reference-positive cells are
\begin{equation}
\mathrm{BS}=\frac{\sum_k w_k(P_k-T_k)^2}{\sum_k w_k},\qquad
\mathrm{BS}_1=\frac{\sum_{k:T_k=1} w_k(P_k-1)^2}{\sum_{k:T_k=1} w_k}.
\end{equation}
$\mathrm{BS}_1$ measures how much probability the ensemble places where the reference is active. For a member mask $M$, with $A(\cdot)$ the area of a set of cells and $d(p,Q)$ the distance in km from cell $p$ to the nearest cell of $Q$,
\begin{equation}
\mathrm{IoU}=\frac{A(T\cap M)}{A(T\cup M)},\qquad
H=\max\Big(\max_{p\in M}d(p,T),\ \max_{p\in T}d(p,M)\Big).
\end{equation}
IoU measures overlap and the Hausdorff distance $H$ the largest displacement between the two masks. The G-score \citep{gilleland_novel_2021} combines the non-overlapping area with the mean distances between the masks: with $n_T$, $n_M$ and $n_{T\cap M}$ the numbers of cells,
\begin{equation}
G=(y_1y_2)^{1/3},\qquad y_1=n_T+n_M-2n_{T\cap M},\qquad y_2=\bar d(T,M)\,n_M+\bar d(M,T)\,n_T,
\end{equation}
where $\bar d(Q,R)$ is the mean over cells of $Q$ of the distance to $R$. IoU, $H$ and $G$ are computed for each member and averaged over members, then storms; higher IoU and lower $\mathrm{BS}$, $\mathrm{BS}_1$, $H$ and $G$ are better.

\paragraph{Joint pressure--wind distributions.}
The gridded-source rows are restricted to forecast times that coincide with an IBTrACS observation containing valid \texttt{USA\_PRES} and \texttt{USA\_WIND}. This mask depends on the reference record, not on successful predicted-track detection. Each available member contributes one pair of regional pressure minimum and wind maximum per retained time; the extrema need not occur at the same grid point. Pressure minima are approximated by the midpoint of the first occupied bin in cached 80-bin pressure histograms, while wind maxima are stored directly. The IBTrACS comparison uses available observations with both quantities for the represented storms, including its finer observation cadence; it is not restricted to the identical 12-hour forecast timestamps. Wind is converted from knots to m/s and pressure to hPa.

Each source's pooled pairs receive equal row weight in a Gaussian KDE (SciPy's default bandwidth rule), rather than equal storm weight. Contours are drawn at the 30th and 70th percentiles of each source's density evaluated at its own points. For the dependence panels, each source's pressures and winds are ranked separately, ties receiving their average rank (ties are frequent because pressure is a histogram-bin midpoint), and rank $r$ among $n$ rows is mapped to $\Phi^{-1}(r/(n+1))$, with $\Phi$ the standard normal distribution function. This keeps each variable's ordering and makes every marginal standard normal, so no increasing transformation of either variable, such as a mean offset, a compressed wind distribution, or a monotone conversion between wind definitions, changes the panel. Spearman's $\rho$ is computed on the same rows; its 95\% interval comes from 200 resamples of storms with replacement, because rows from one storm are not independent. Source-specific missing data, temporal weighting, and the pressure-bin approximation must still be kept in view.

\paragraph{Data-volume estimate.}
For scale, seven global fields sampled every 12 hours on a $0.25^\circ$ grid correspond to approximately 23~TB of uncompressed float32 data.\footnote{This is a data-volume estimate, not a measurement of disk usage: $26{,}700$ member forecasts (178 storms $\times$ 50 members $\times$ 3 models) $\times\ 30$ forecast times $\times 721 \times 1440$ grid points $\times 7$ fields $\times 4$ bytes, with $1\,\mathrm{TB}=10^{12}$ bytes. It excludes additional variables, finer temporal output, reference fields, and intermediate files.}

\section{Additional results}
\label{app:additional}

\subsection{Stability across storm subgroups}
\label{app:stability}

\begin{figure}[htbp]
  \centering
  \includegraphics[width=\linewidth]{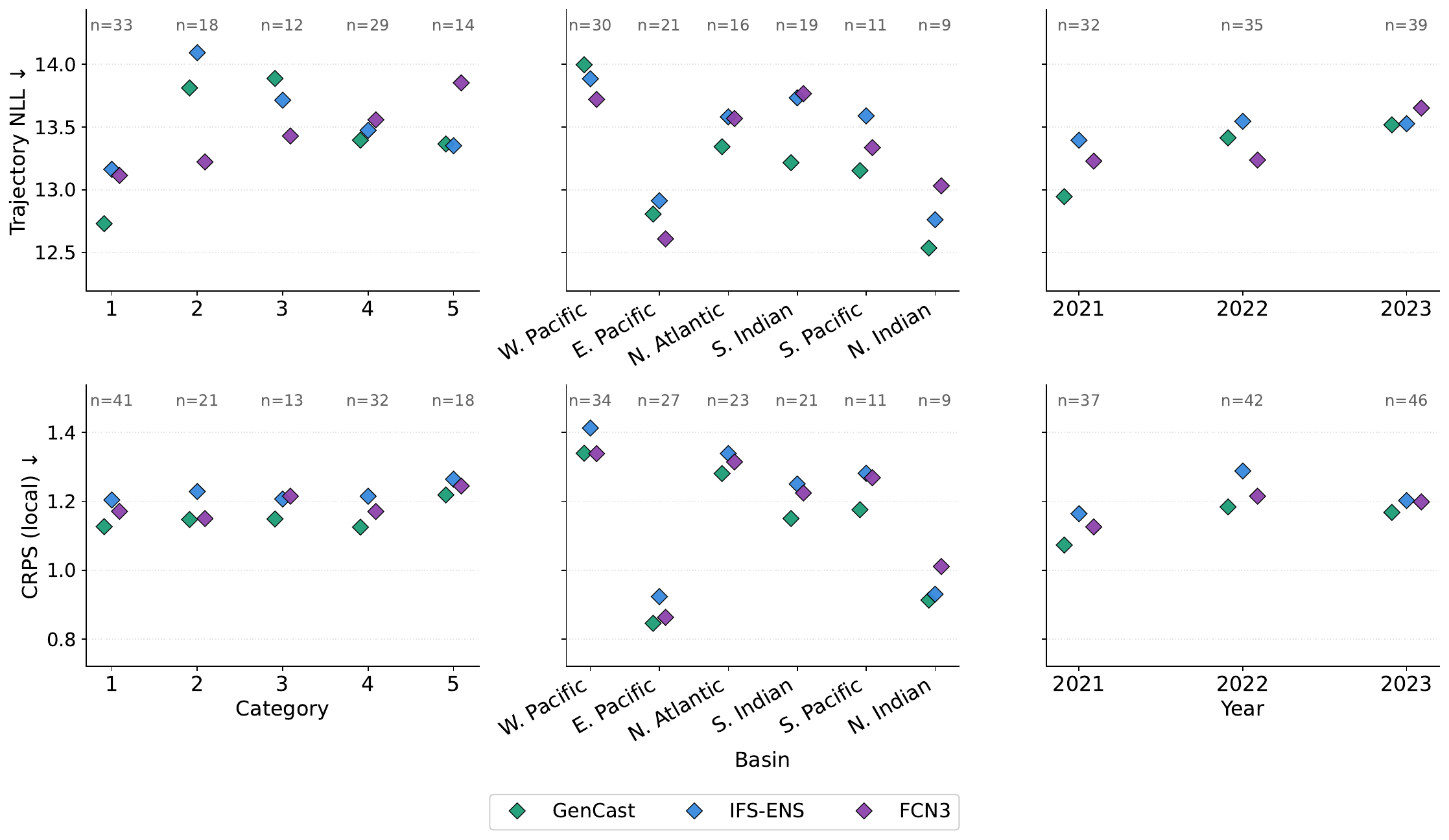}
  \caption{\textbf{Stability of the model comparison across subgroups (2021--2023).} Track NLL (top row) and cyclone-local CRPS (bottom row) for GenCast, IFS-ENS and FCN3, averaged within subgroups defined by Saffir--Simpson category (left), ocean basin (center) and year (right); $n$ gives the subgroup size. The NLL uses the 106 retained storms and the CRPS all 125, hence the different subgroup sizes. The three models move together across nearly all subgroups.}
  \label{fig:stability-forest-fcn3}
\end{figure}

\begin{figure}[htbp]
  \centering
  \includegraphics[width=0.664\linewidth]{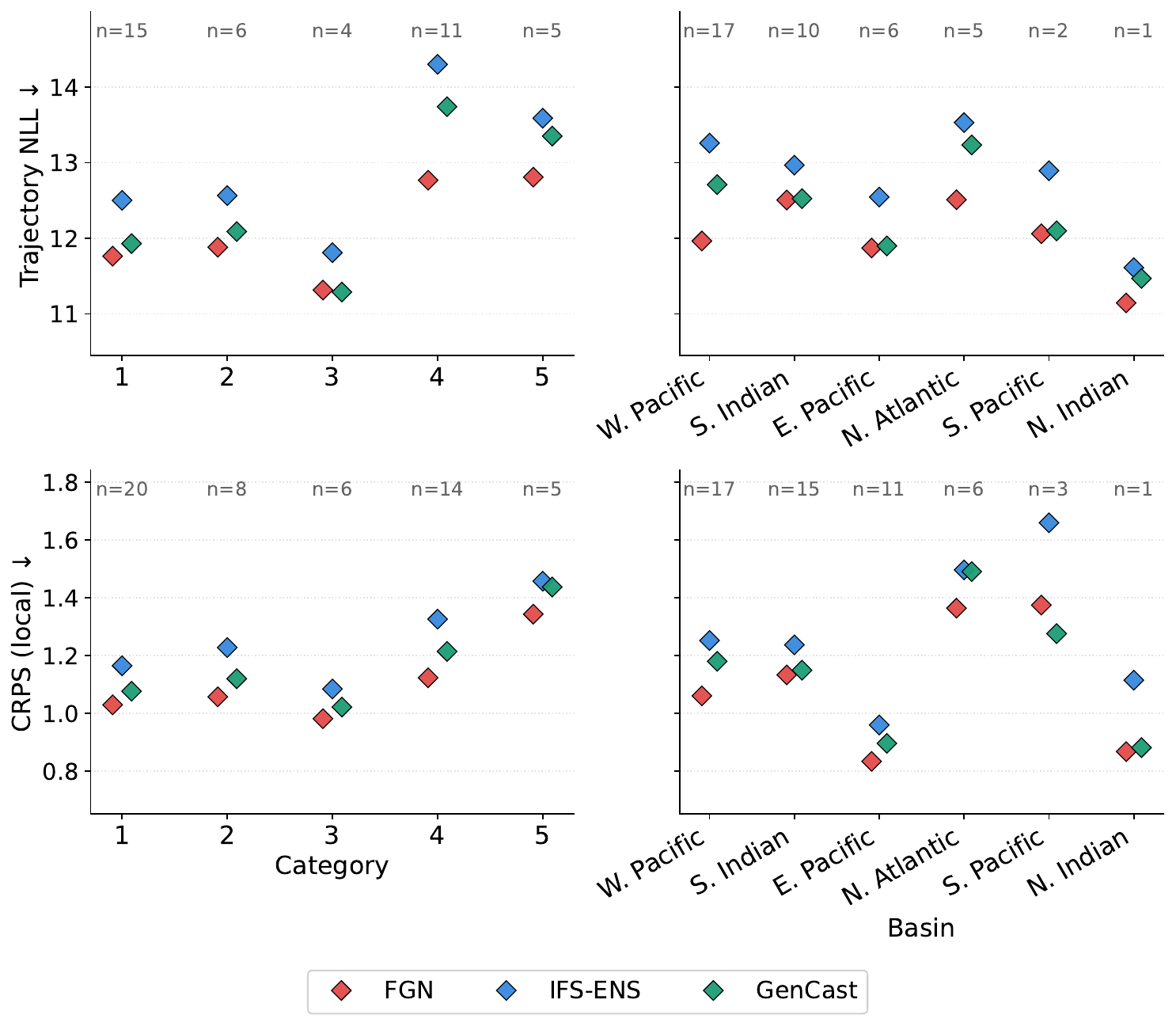}
  \caption{\textbf{Stability of the model comparison across subgroups (2025).} As in Figure~\ref{fig:stability-forest-fcn3}, for FGN, IFS-ENS and operational GenCast, at the same scale; subgroups are defined by Saffir--Simpson category (left) and ocean basin (right). The NLL uses the 41 retained storms and the CRPS all 53. Some basins contain very few storms, down to $n{=}1$, which explains the larger gaps there.}
  \label{fig:stability-forest-fgn}
\end{figure}

To check that the comparisons of Sections~\ref{sec:evaluation} and~\ref{sec:operational} are not driven by a few storms or by the mix of storms in each cohort, we average the track NLL and the cyclone-local CRPS within subgroups defined by Saffir--Simpson category, ocean basin and, for 2021--2023, year (Figures~\ref{fig:stability-forest-fcn3} and~\ref{fig:stability-forest-fgn}). IFS-ENS is a numerical model, so its variation across subgroups serves as a reference for how difficult the storms in each subgroup are: if the AI models vary in the same way, their variation reflects the storms rather than the model.

In both cohorts the AI models mostly follow the IFS-ENS pattern across subgroups. Larger gaps occur in the smallest subgroups, down to a single storm in the 2025 basin breakdown. In 2025, IFS-ENS has the highest NLL and CRPS in every subgroup, and FGN the lowest in all but two: GenCast is marginally ahead on NLL in category 3 and on CRPS in the South Pacific, which has three storms.

The one clear exception is the 2021--2023 track NLL by category. GenCast still varies with IFS-ENS, but FCN3 does not: its NLL increases steadily with category, whereas IFS-ENS peaks at category 2 and decreases thereafter, so the ranking of the three models changes from one category to the next. In the year breakdown, the 2023 IFS-ENS forecasts mix cycles 47r3 and 48r1 (Table~\ref{tab:ifs-cycles}).

\subsection{Wind--pressure relationship}
\label{app:wind_pressure}

\begin{figure}[htbp]
    \centering
    \begin{subfigure}[t]{0.49\linewidth}
        \centering
        \includegraphics[width=\linewidth]{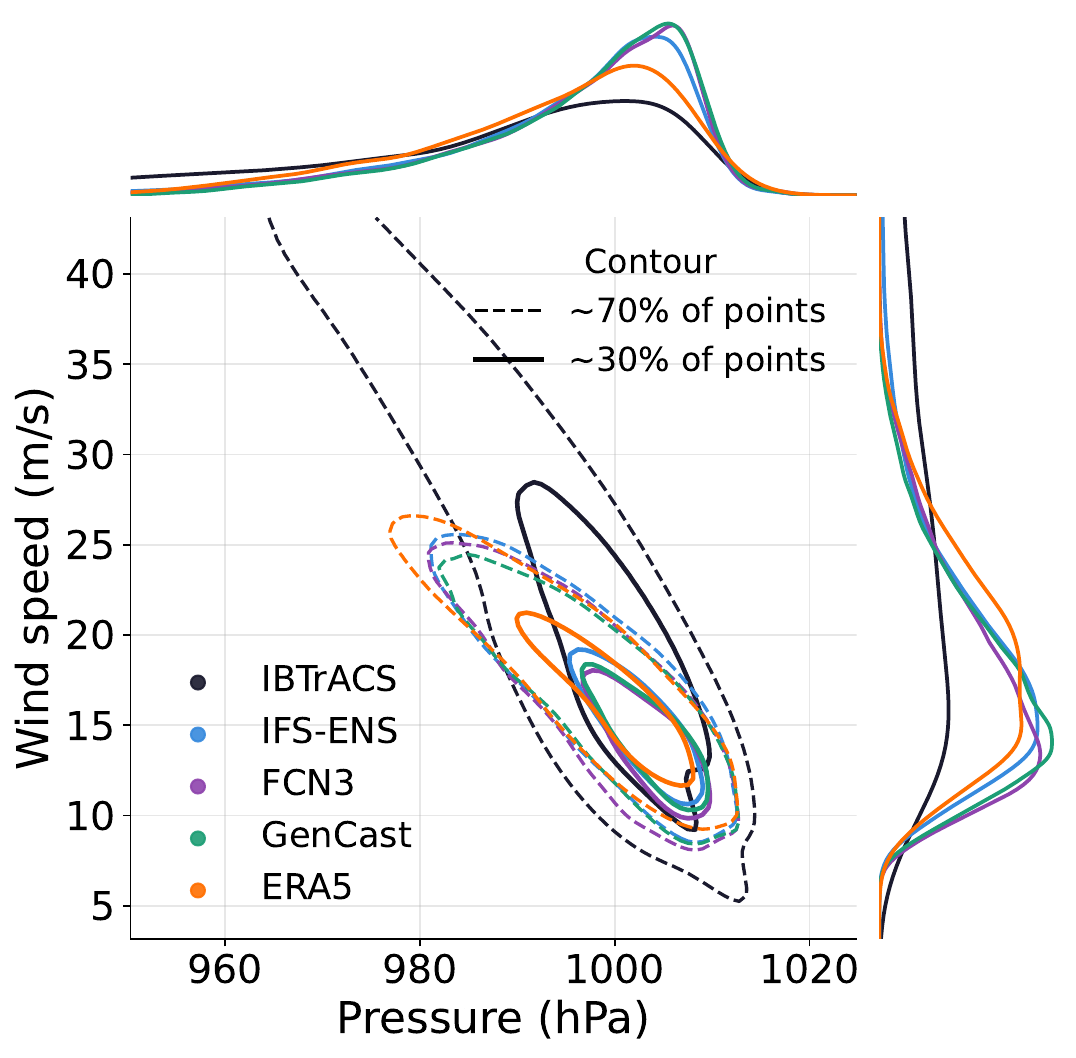}
        \caption{Joint wind--pressure distributions and their marginals.}
    \end{subfigure}\hfill
    \begin{subfigure}[t]{0.49\linewidth}
        \centering
        \includegraphics[width=\linewidth]{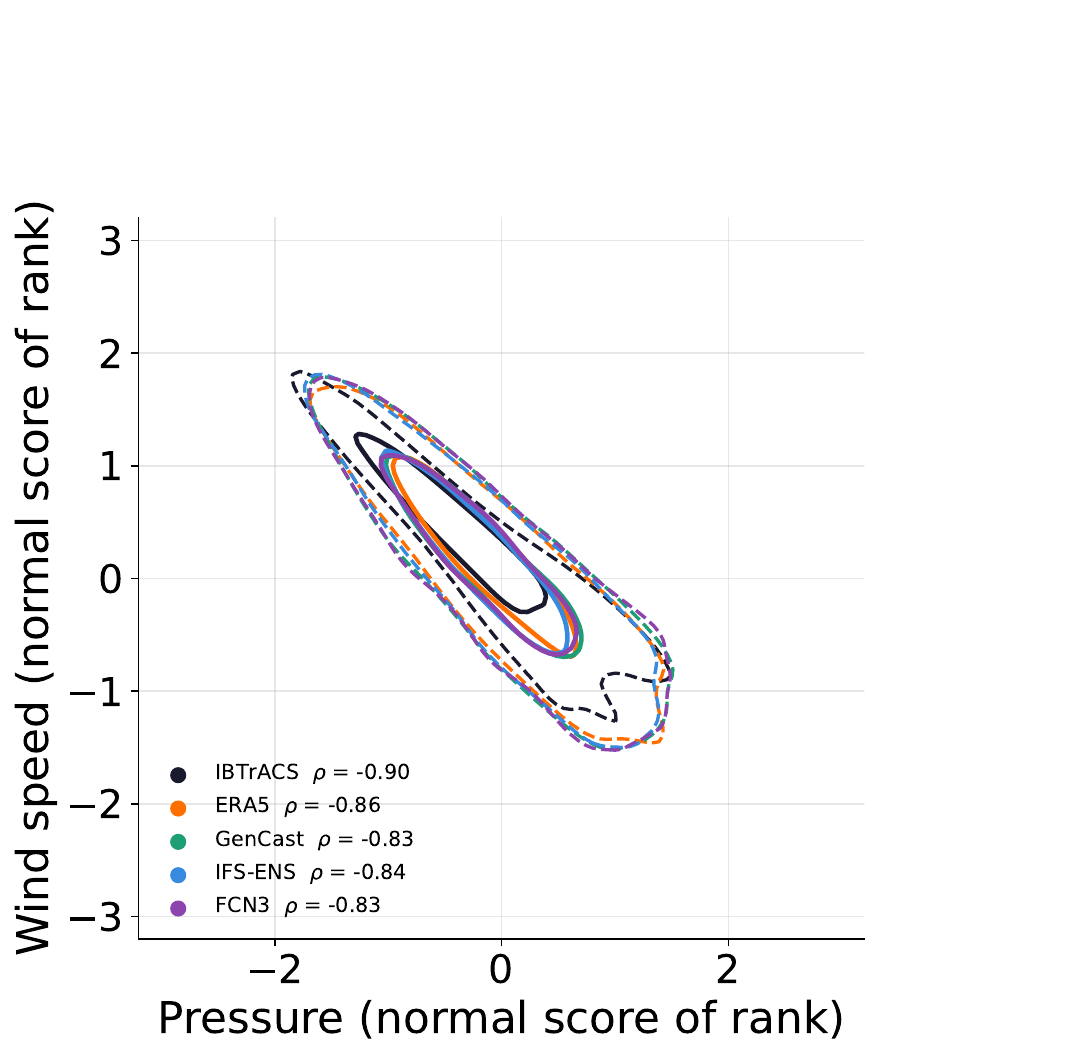}
        \caption{Rank dependence (copula), marginal biases removed; Spearman's $\rho$ in the legend.}
    \end{subfigure}
    \caption{\textbf{Wind--pressure distributions in the ERA5 setting (2021--2023).} Same construction as Figure~\ref{fig:v2_joint_operational}. All three models have nearly the same dependence as ERA5 (b): $\rho$ is $-0.83$ for FCN3 and GenCast, $-0.84$ for IFS-ENS and $-0.86$ for ERA5, with overlapping 95\% intervals. IBTrACS uses sustained rather than gridded instantaneous winds.}
    \label{fig:v2_joint_era5}
\end{figure}

Figure~\ref{fig:v2_joint_era5} repeats the analysis of Section~\ref{sec:storm_structure} for the 2021--2023 setting (GenCast, FCN3, and IFS-ENS), with ERA5 as the gridded reference. The three models have nearly the same joint distribution; compared with ERA5, their pressures are slightly shallower and their winds slightly weaker (a). Their dependence is close to ERA5's (b). As in the operational setting, the models differ from the reference in intensity rather than in how wind varies with pressure.

\subsection{Rank histograms}
\label{app:rank_histograms}

\begin{figure}[htbp]
    \centering
    \begin{subfigure}[t]{0.49\linewidth}
        \centering
        \includegraphics[width=\linewidth]{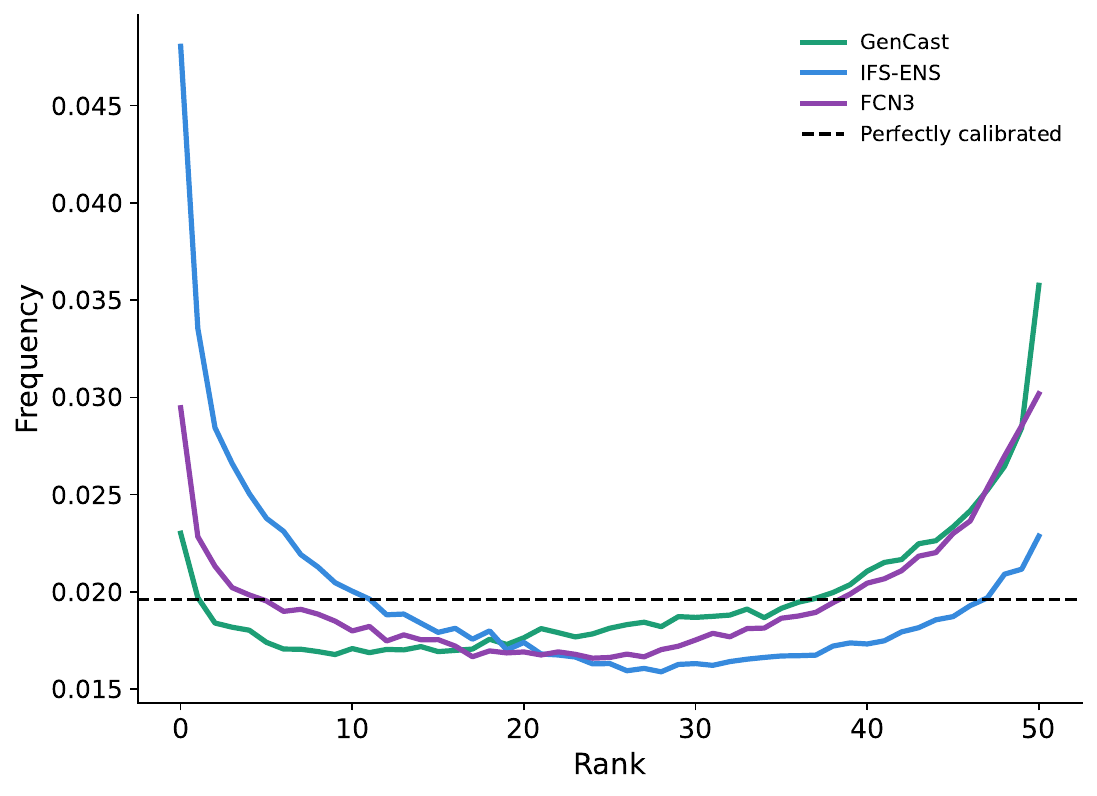}
        \caption{2021--2023, ERA5 reference.}
    \end{subfigure}\hfill
    \begin{subfigure}[t]{0.49\linewidth}
        \centering
        \includegraphics[width=\linewidth]{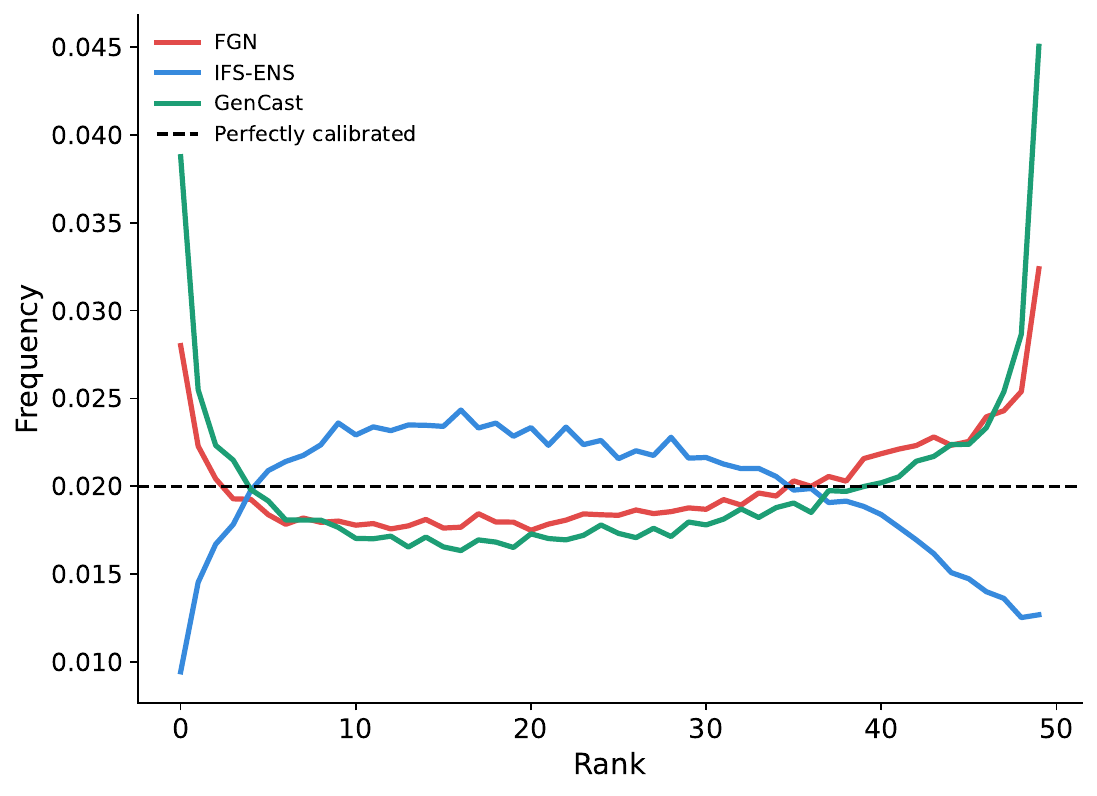}
        \caption{2025, HRES-fc0 reference.}
    \end{subfigure}
    \caption{\textbf{Rank histograms of 10m wind footprints.} At each grid point of a storm's region, the reference footprint (maximum wind over the forecast window) is ranked within the members' footprints; frequencies are averaged across storms. The dashed line is the uniform frequency of a calibrated ensemble.}
    \label{fig:rank-footprint}
\end{figure}

\begin{figure}[htbp]
    \centering
    \begin{subfigure}[t]{\linewidth}
        \centering
        \includegraphics[width=\linewidth]{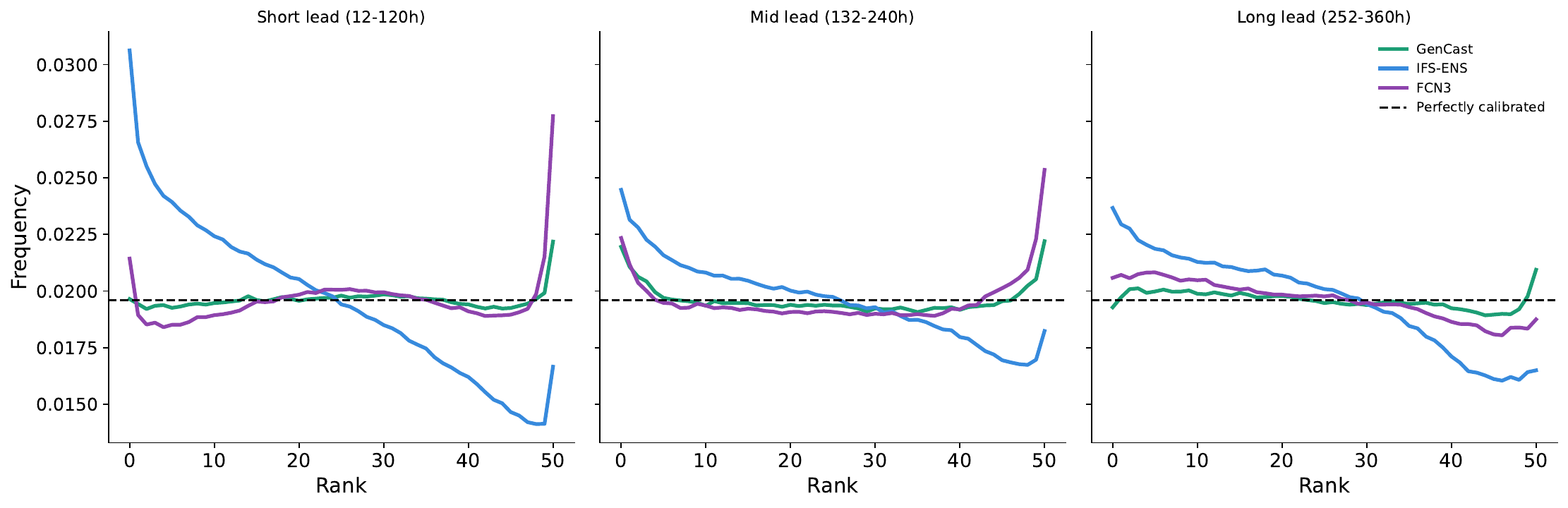}
        \caption{2021--2023, ERA5 reference.}
    \end{subfigure}

    \vspace{0.5em}
    \begin{subfigure}[t]{\linewidth}
        \centering
        \includegraphics[width=\linewidth]{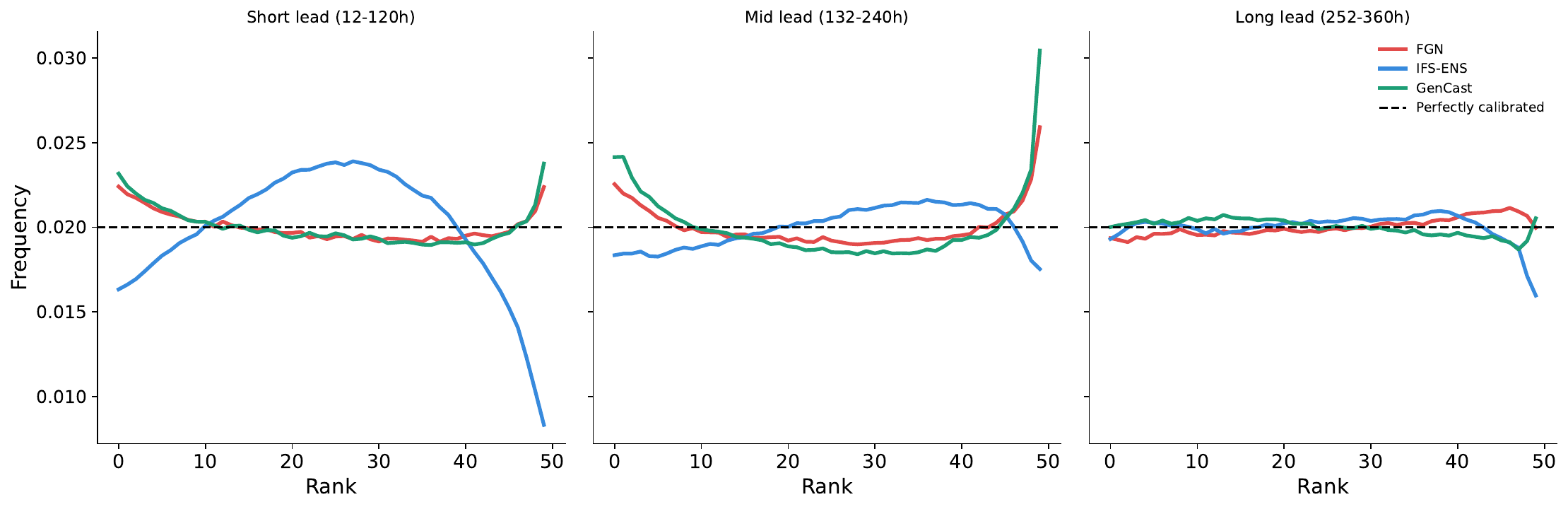}
        \caption{2025, HRES-fc0 reference.}
    \end{subfigure}
    \caption{\textbf{Wind-field calibration by lead-time range.} At each grid point and lead time, the reference 10m wind speed is ranked within the forecast ensemble. Rank frequencies are averaged across storms and within the displayed lead-time ranges; dashed lines indicate uniform ranks. In 2025 (b), the AI ensembles show excesses at both ends at short and intermediate leads, consistent with insufficient spread; the larger upper-rank excess at intermediate leads is also consistent with weak predicted winds. Long-lead histograms are closer to uniform.}
    \label{fig:v2_operational_calibration}
\end{figure}

\begin{figure}[htbp]
    \centering
    \includegraphics[width=\linewidth]{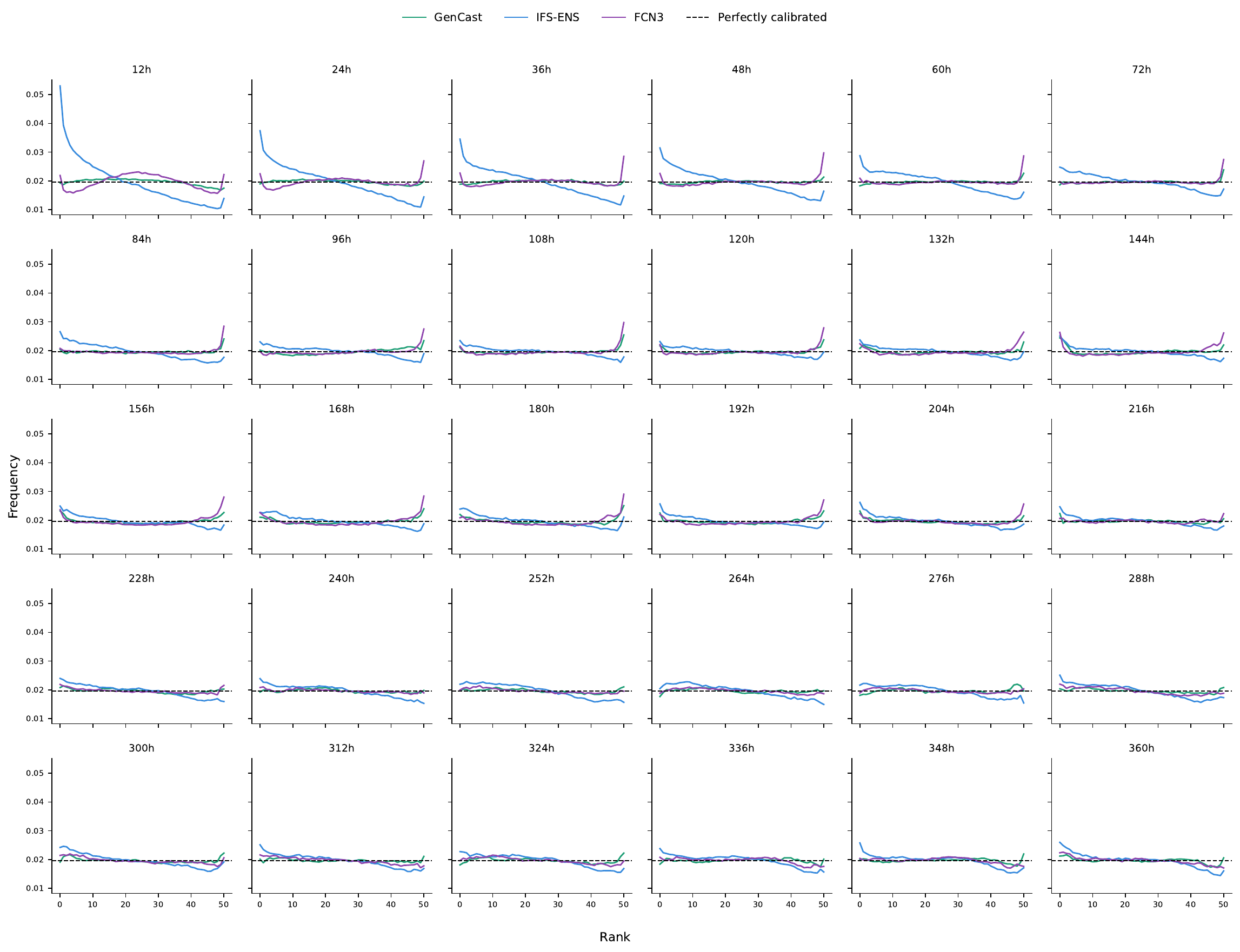}
    \caption{\textbf{Wind-field rank histograms at each lead time (2021--2023, ERA5 reference).} As in Figure~\ref{fig:v2_operational_calibration}a, with one panel per 12-hour lead time instead of three ranges.}
    \label{fig:rank-leadtime-fcn3}
\end{figure}

\begin{figure}[htbp]
    \centering
    \includegraphics[width=\linewidth]{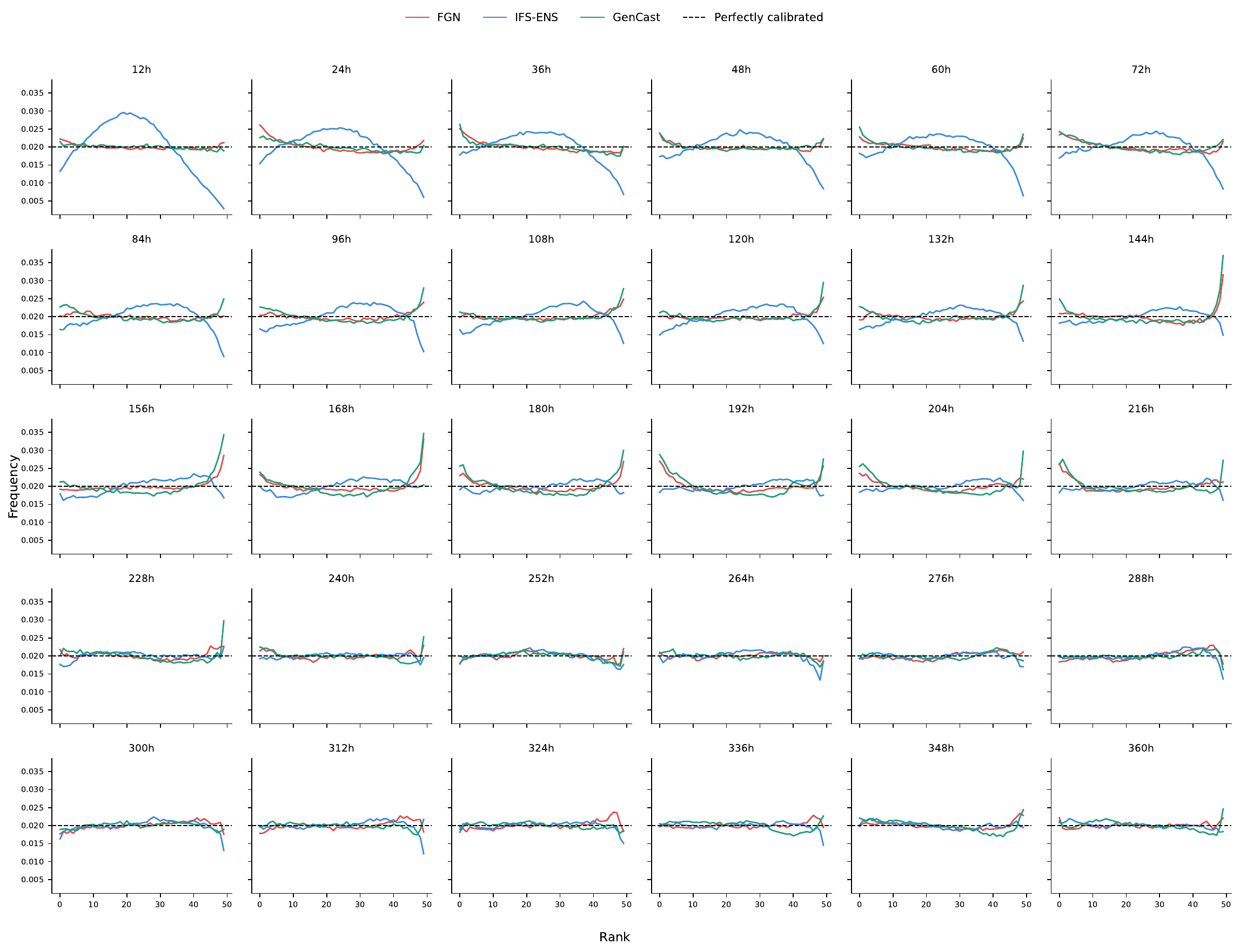}
    \caption{\textbf{Wind-field rank histograms at each lead time (2025, HRES-fc0 reference).} As in Figure~\ref{fig:v2_operational_calibration}b, with one panel per 12-hour lead time instead of three ranges.}
    \label{fig:rank-leadtime-fgn}
\end{figure}

A rank histogram counts, at each grid point, how many ensemble members fall below the reference value. For a calibrated ensemble, the reference is equally likely to take any of the $N+1$ ranks and the histogram is flat. A U shape, with the reference too often below or above all members, indicates insufficient spread; a $\bigcap$ shape, with the reference too often near the middle of the ensemble, indicates excessive spread; an asymmetry indicates a bias. A high last rank means the reference is often stronger than every member, i.e.\ the forecast winds are too weak; a high first rank means the opposite. We show 10m wind speed, for storm footprints (Figure~\ref{fig:rank-footprint}), for forecast fields grouped into three lead-time ranges (Figure~\ref{fig:v2_operational_calibration}), and for each lead time (Figures~\ref{fig:rank-leadtime-fcn3} and~\ref{fig:rank-leadtime-fgn}).

In 2021--2023, against ERA5, IFS-ENS places the reference below all its members too often, at every lead time and on footprints: its winds are stronger than ERA5's, consistent with ERA5's underestimation of peak winds \citep{dulac_assessing_2024} rather than an IFS-ENS error. Both AI models have U-shaped footprint histograms, indicating too little spread, and GenCast's larger excess at the last rank shows footprint winds weaker than ERA5's. By lead time, the AI models underestimate the wind at short leads, and FCN3 also at intermediate leads; at long leads their histograms are close to uniform.

In 2025, against HRES-fc0, FGN and GenCast have U-shaped histograms on footprints and at intermediate leads, indicating too little spread, with a larger excess at the last rank, i.e.\ some underestimation of the wind. IFS-ENS has a $\bigcap$ shape on footprints and at short and intermediate leads, indicating too much spread. At long leads, all three histograms are close to uniform, but most storms have faded out by then.

\subsection{Correlation between track and intensity errors}
\label{app:track_intensity}

\begin{figure}[htbp]
  \centering
  \includegraphics[width=\linewidth]{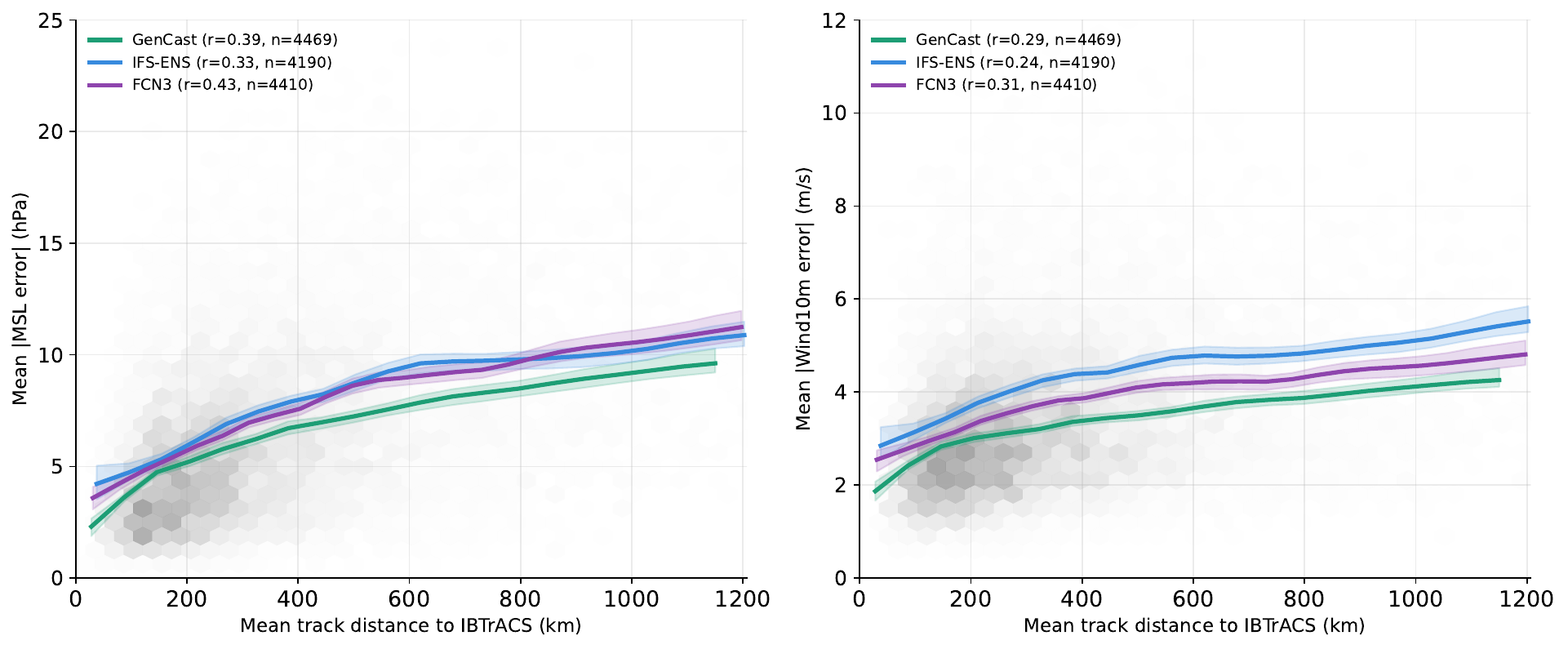}
  \caption{\textbf{Track error against intensity error, by model (2021--2023, ERA5 reference).} Each matched ensemble member contributes its mean track distance to IBTrACS and its mean absolute intensity error, for mean sea-level pressure (left) and 10m wind speed (right). Lines show a LOWESS trend per model with a 95\% bootstrap band, over the hexagonal density of the members; the Pearson correlation $r$ and the number of members $n$ are given per model.}
  \label{fig:loc-int-corr}
\end{figure}

\begin{figure}[htbp]
  \centering
  \includegraphics[width=\linewidth]{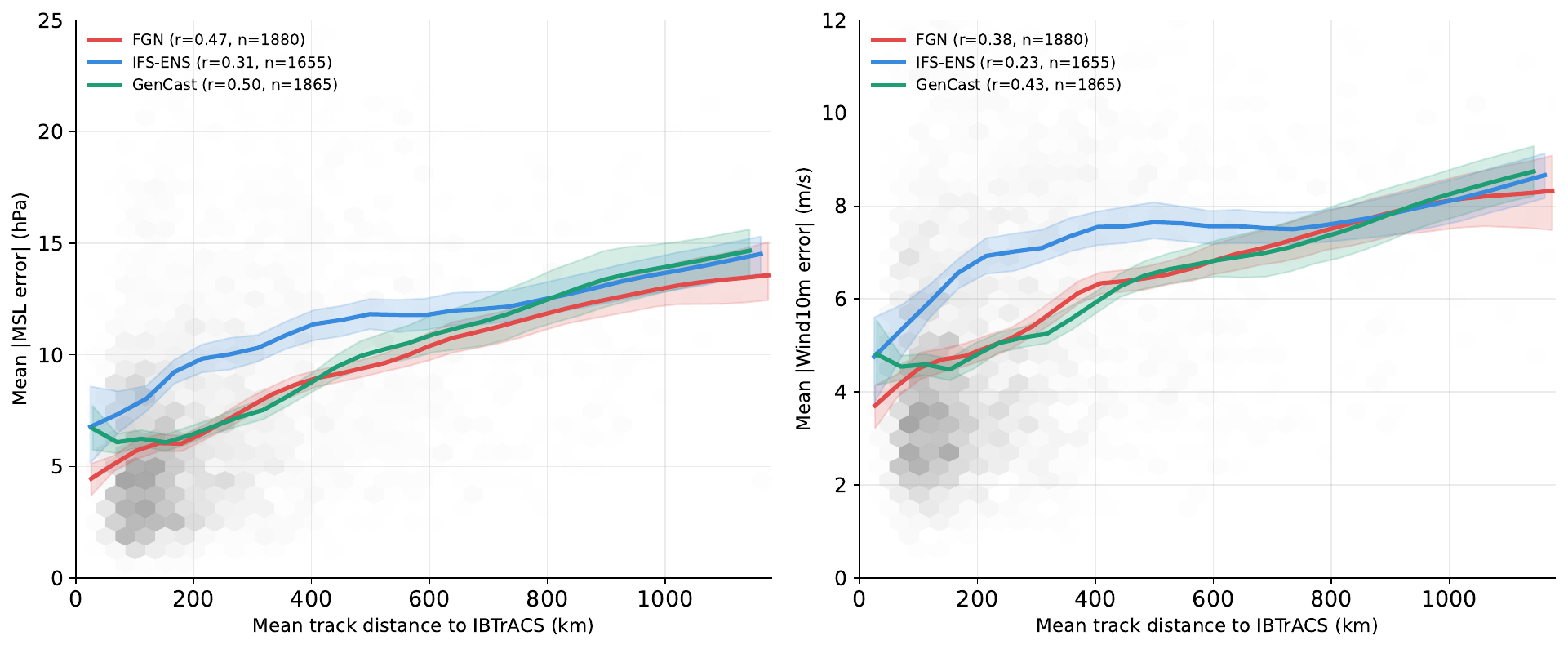}
  \caption{\textbf{Track error against intensity error, by model (2025, HRES-fc0 reference).} As in Figure~\ref{fig:loc-int-corr}, for FGN, IFS-ENS and operational GenCast.}
  \label{fig:loc-int-corr-fgn}
\end{figure}

We check whether a member that misplaces a storm also misjudges its intensity. For each matched ensemble member of the 106 (2021--2023) and 41 (2025) retained storms, we average over its forecast times the distance between its storm centre and the IBTrACS position, and the absolute error of its regional pressure minimum and 10m wind maximum against the gridded reference (ERA5 for 2021--2023, HRES-fc0 for 2025). This gives one point per storm and member. Figures~\ref{fig:loc-int-corr} and~\ref{fig:loc-int-corr-fgn} show a LOWESS trend per model with a 95\% bootstrap band, and the Pearson correlation $r$. The LOWESS trend is a local linear regression: at each track error, a straight line is fitted by weighted least squares to the 30\% of members closest to it, with weights decreasing with distance \citep{cleveland_robust_1979}. The band gives the 2.5th and 97.5th percentiles of the trend over 150 bootstrap resamples of the members.

Track and intensity errors are linked: members that place the storm further from its observed position also tend to have larger intensity errors, in both settings and for both variables. The correlation remains moderate, however, and members with similar track errors span a wide range of intensity errors. An accurate track therefore does not imply an accurate intensity, which is why we evaluate the two separately. IFS-ENS is the least correlated model in every case: $r=0.33$ for pressure and $0.24$ for wind, against $0.39$--$0.43$ and $0.29$--$0.31$ for the AI models in 2021--2023, and $0.31$ and $0.23$ against $0.47$--$0.50$ and $0.38$--$0.43$ in 2025.

\subsection{Spatial agreement}
\label{app:operational_diagnostics}

These diagnostics complement the spatial comparison of Section~\ref{sec:operational} (Figure~\ref{fig:v2_location_lead}). They use HRES-fc0 as the field reference and the 53 storms with scores for all three models. Each field is turned into a mask with a percentile threshold, the reference using its own and all members of a model sharing one (Appendix~\ref{app:reproducibility}), which removes sensitivity to uniform intensity shifts while retaining errors in spatial placement and shape.

\begin{figure}[htbp]
    \centering
    \includegraphics[width=\linewidth]{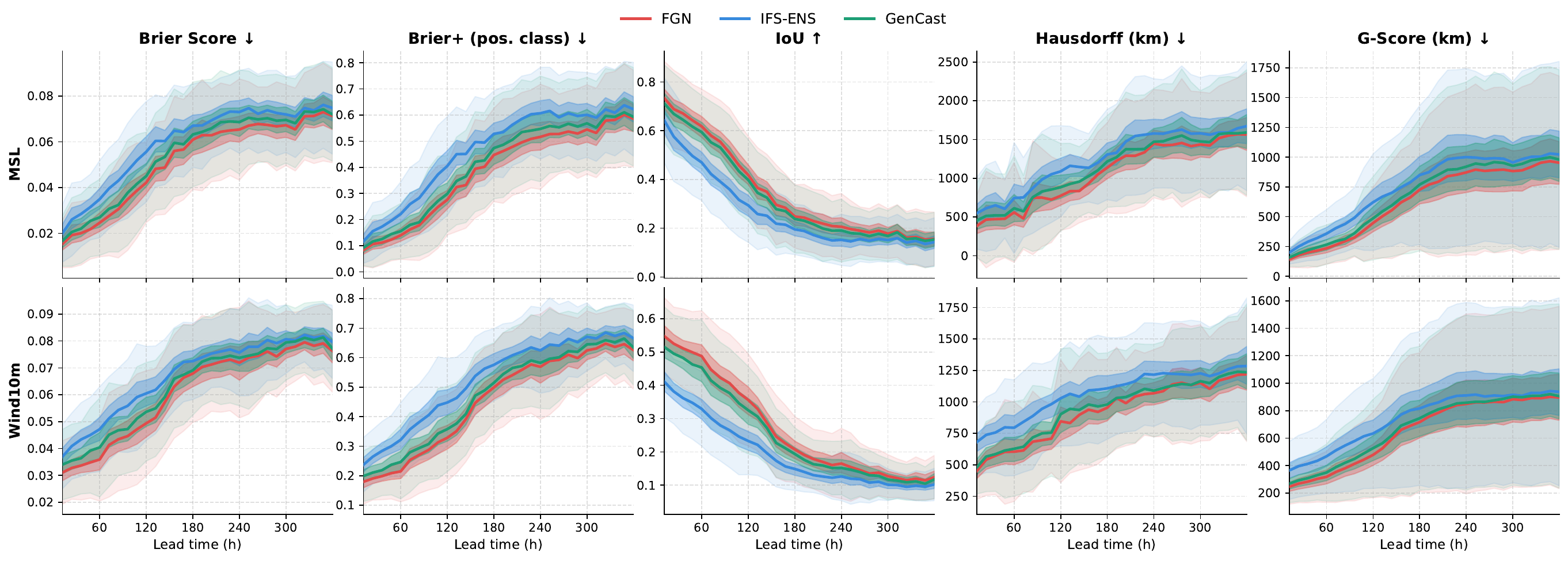}
    \caption{\textbf{Location metrics by lead time (2025, HRES-fc0 reference).} Columns: Brier score, positive-class Brier score, IoU, Hausdorff distance and G-score; arrows indicate whether higher or lower is better. Rows: sea-level pressure and 10m wind speed, at the 90th percentile. All members of a model share one threshold; the reference uses its own. Lines are means over the 53 storms; dark bands are 95\% bootstrap confidence intervals of the mean and light bands $\pm1$ standard deviation across storms.}
    \label{fig:location-lead-full}
\end{figure}

\begin{figure}[htbp]
    \centering
    \includegraphics[width=\linewidth]{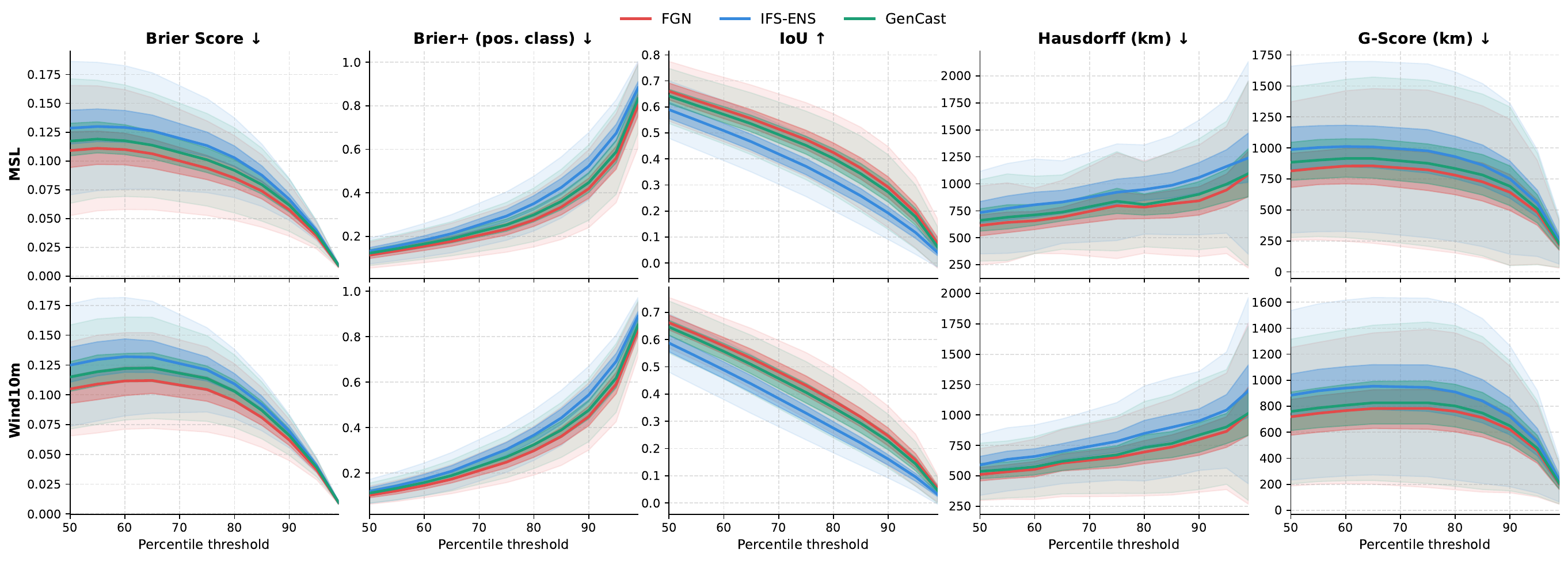}
    \caption{\textbf{Location metrics on storm footprints (2025, HRES-fc0 reference).} As in Figure~\ref{fig:location-lead-full}, for footprints (maximum wind or minimum pressure over the forecast window) as a function of the percentile threshold.}
    \label{fig:location-footprint}
\end{figure}

Figure~\ref{fig:location-lead-full} extends Figure~\ref{fig:v2_location_lead} to sea-level pressure and to all five metrics, by lead time at the 90th percentile, and Figure~\ref{fig:location-footprint} applies the same metrics to footprints, the maximum wind or minimum pressure over the forecast window, across percentile thresholds from 50 to 99. Both give the same result as the tracking scores: IFS-ENS scores worst on every metric and for both variables, with the largest gaps before about 180 hours, and FGN scores best, closely followed by GenCast.

\subsection{Energy spectra}
\label{app:spectra}

\begin{figure}[htbp]
    \centering
    \begin{subfigure}[t]{\linewidth}
        \centering
        \includegraphics[width=\linewidth]{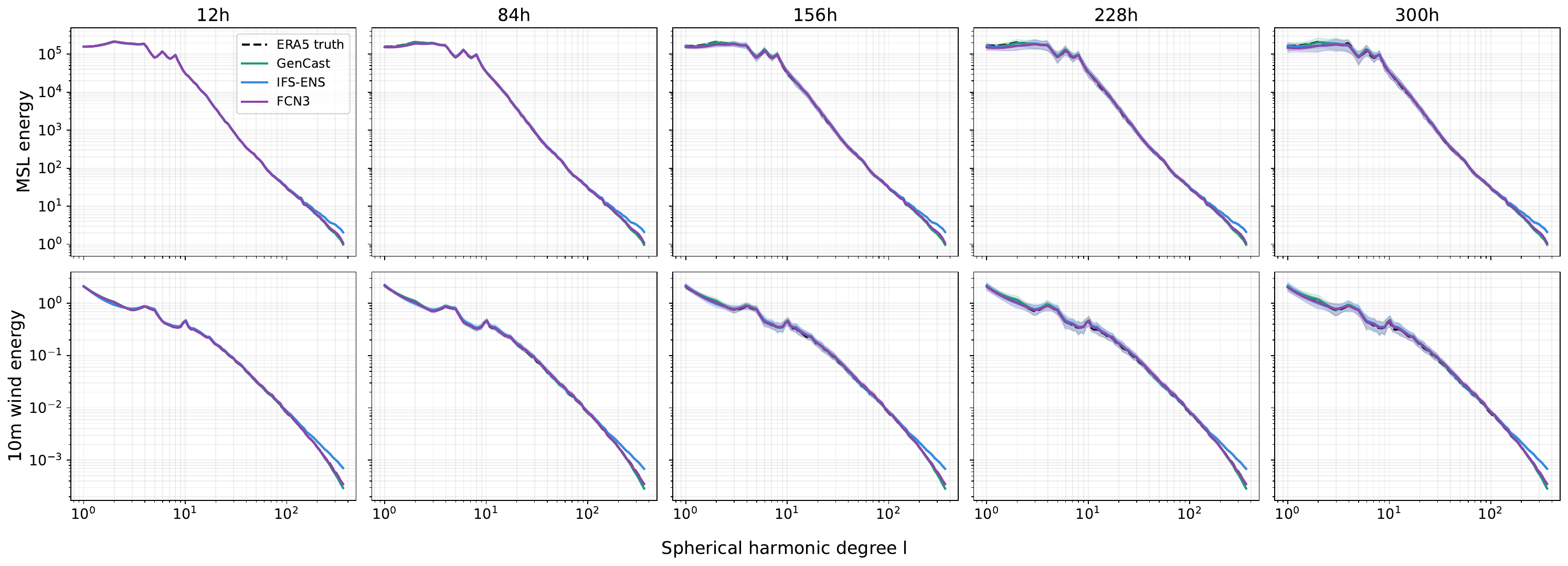}
        \caption{All degrees.}
    \end{subfigure}

    \vspace{0.5em}
    \begin{subfigure}[t]{\linewidth}
        \centering
        \includegraphics[width=\linewidth]{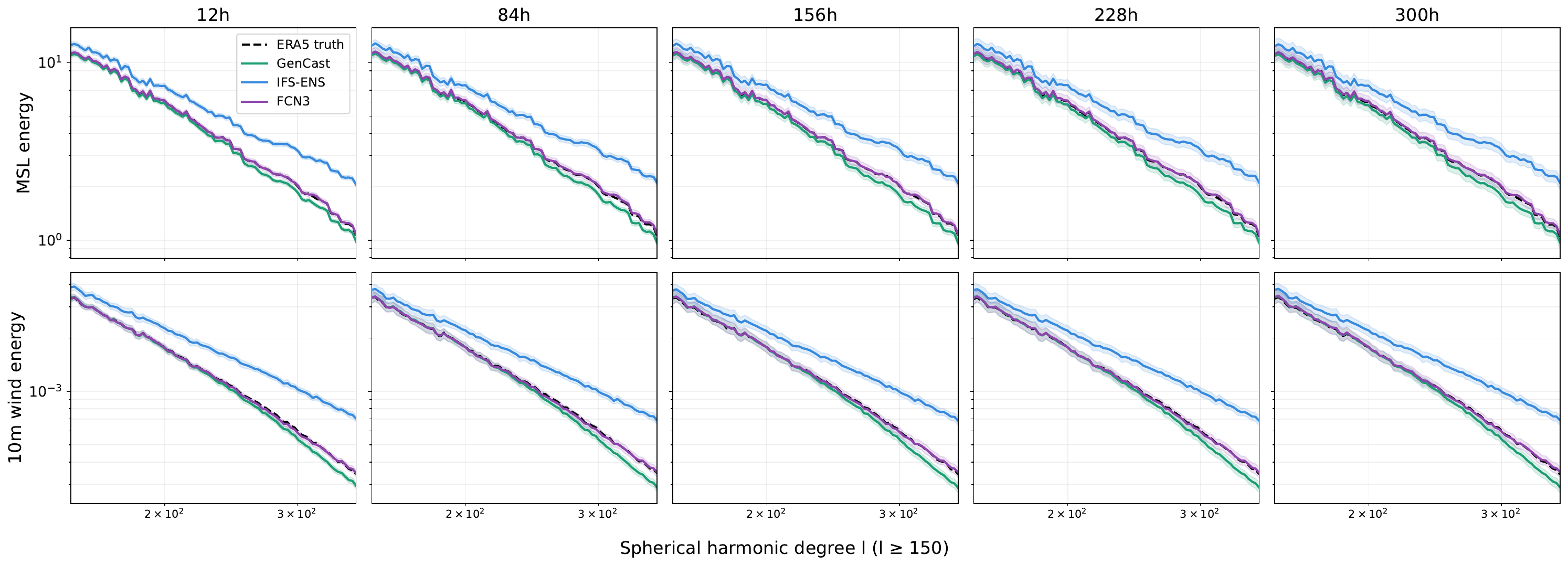}
        \caption{Degrees $l \geq 150$.}
    \end{subfigure}
    \caption{\textbf{Energy spectra of global sea-level pressure (top rows) and 10m wind speed (bottom rows) by lead time (2021--2023, ERA5 reference).} Lines are the members' spectra averaged over members and storms, and bands $\pm1$ standard deviation across members; the dashed line is ERA5.}
    \label{fig:spectrum-fcn3}
\end{figure}

\begin{figure}[htbp]
    \centering
    \begin{subfigure}[t]{\linewidth}
        \centering
        \includegraphics[width=\linewidth]{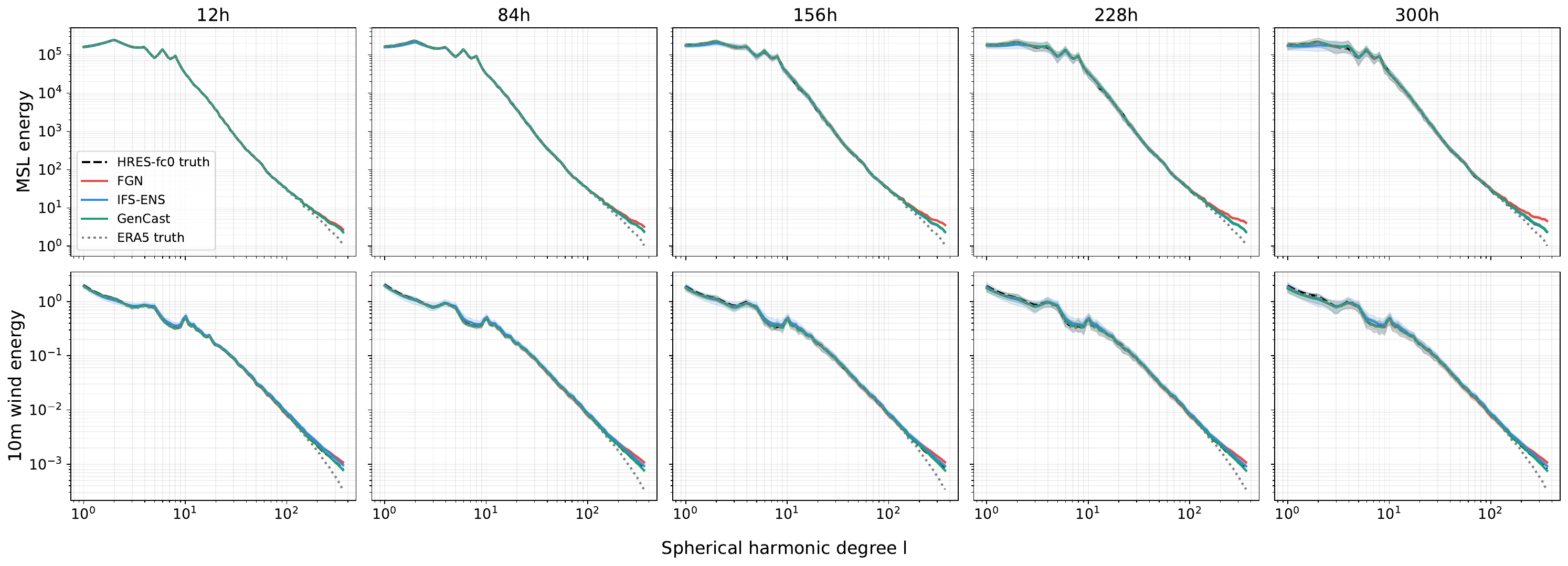}
        \caption{All degrees.}
    \end{subfigure}

    \vspace{0.5em}
    \begin{subfigure}[t]{\linewidth}
        \centering
        \includegraphics[width=\linewidth]{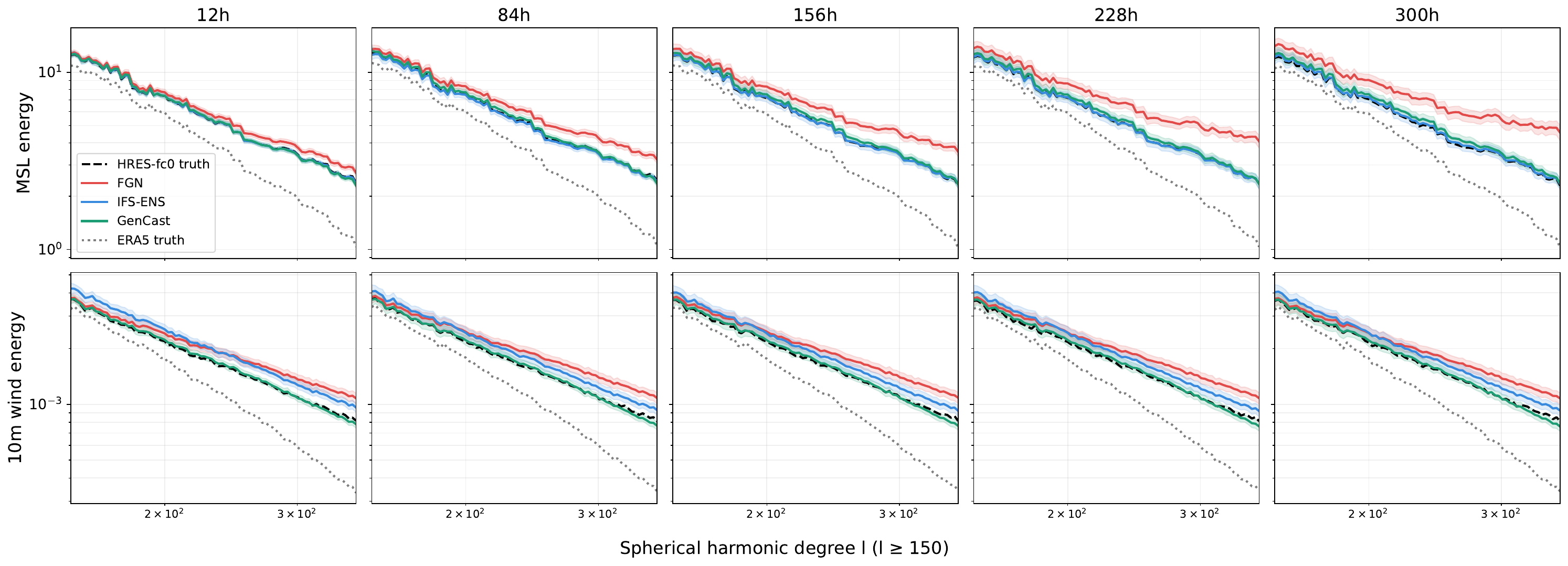}
        \caption{Degrees $l \geq 150$.}
    \end{subfigure}
    \caption{\textbf{Energy spectra (2025, HRES-fc0 reference).} As in Figure~\ref{fig:spectrum-fcn3}, for FGN, IFS-ENS and operational GenCast; the dashed line is HRES-fc0 and the dotted line ERA5.}
    \label{fig:spectrum-fgn}
\end{figure}

We compute the power spectrum of the global sea-level pressure and 10m wind fields as a function of spherical harmonic degree $l$, at five lead times, and average it over storms (Figures~\ref{fig:spectrum-fcn3} and~\ref{fig:spectrum-fgn}). This checks whether a model keeps the distribution of energy across spatial scales during the rollout rather than smoothing out small scales. At large scales all models follow the reference; they separate only at the smallest scales, shown from $l=150$ in panels (b).

In 2021--2023, FCN3 follows the ERA5 spectrum at all degrees and lead times, and GenCast has slightly less small-scale energy than ERA5. IFS-ENS has markedly more, at every lead time including 12 hours, consistent with its native grid being finer than the $0.25^\circ$ evaluation grid (Table~\ref{tab:ifs-cycles}). In 2025, HRES-fc0 itself has much more small-scale energy than ERA5, and the operational models follow HRES-fc0 rather than ERA5 at these scales. GenCast stays close to HRES-fc0 for both variables, slightly below it for wind. IFS-ENS matches HRES-fc0 for pressure and has more energy for wind. FGN has more small-scale energy than HRES-fc0 for both variables; for pressure the excess is small at 12 hours and grows with lead time.

\end{document}